\documentclass[letterpaper]{article} 
\usepackage{aaai25}  
\usepackage{times}  
\usepackage{helvet}  
\usepackage{courier}  
\usepackage[hyphens]{url}  
\usepackage{graphicx} 
\usepackage{natbib}  
\usepackage{caption} 
\usepackage{algorithm}
\usepackage{algorithmic}
\usepackage{multirow}
\usepackage{color}
\usepackage{booktabs}
\usepackage{amssymb}
\usepackage{amsmath}
\usepackage{float}
\usepackage{subfig}

\usepackage{newfloat}
\usepackage{listings}
\DeclareCaptionStyle{ruled}{labelfont=normalfont,labelsep=colon,strut=off} 
\floatstyle{ruled}
\newfloat{listing}{tb}{lst}{}
\floatname{listing}{Listing}
\title{Mono2Stereo: Monocular Knowledge Transfer for Enhanced Stereo Matching}
\author {
    Yuran Wang\footnotemark[1]\textsuperscript{\rm 1},
    Yingping Liang\footnotemark[1]\textsuperscript{\rm 1},
    Hesong Li\textsuperscript{\rm 1},
    Ying Fu\footnotemark[2]\textsuperscript{\rm 1}
}
\affiliations {
    \textsuperscript{\rm 1}Beijing Institute of Technology
}

\usepackage{bibentry}

\begin{document}

\maketitle

\renewcommand{\thefootnote}{\fnsymbol{footnote}} 
\footnotetext[1]{These authors contributed equally to this work.} 
\footnotetext[2]{Corresponding authors.} 

\begin{abstract}
The generalization and performance of stereo matching networks are limited due to the domain gap of the existing synthetic datasets and the sparseness of GT labels in the real datasets. In contrast, monocular depth estimation has achieved significant advancements, benefiting from large-scale depth datasets and self-supervised strategies. To bridge the performance gap between monocular depth estimation and stereo matching, we propose leveraging monocular knowledge transfer to enhance stereo matching, namely \textbf{Mono2Stereo}. We introduce knowledge transfer with a two-stage training process, comprising synthetic data pre-training and real-world data fine-tuning. In the pre-training stage, we design a data generation pipeline that synthesizes stereo training data from monocular images. This pipeline utilizes monocular depth for warping and novel view synthesis and employs our proposed Edge-Aware (EA) inpainting module to fill in missing contents in the generated images. In the fine-tuning stage, we introduce a Sparse-to-Dense Knowledge Distillation (S2DKD) strategy encouraging the distributions of predictions to align with dense monocular depths. This strategy mitigates issues with edge blurring in sparse real-world labels and enhances overall consistency. Experimental results demonstrate that our pre-trained model exhibits strong zero-shot generalization capabilities. Furthermore, domain-specific fine-tuning using our pre-trained model and S2DKD strategy significantly increments in-domain performance. The code will be made available soon.
\end{abstract}

%

\begin{figure}[ht]
    \centering
    \includegraphics[width=0.95 \linewidth]{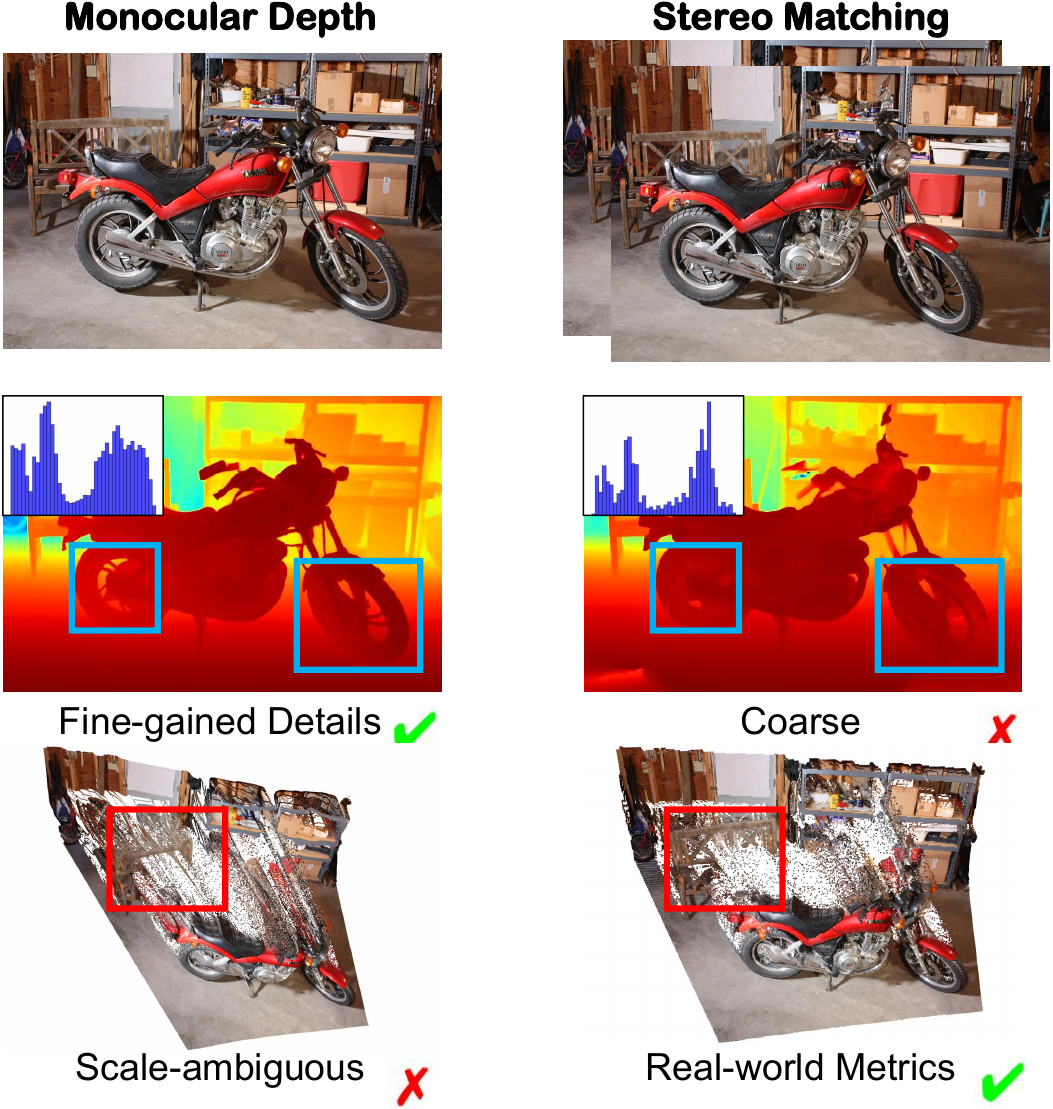}
    \caption{Comparison between Monocular Depth and Stereo Matching for depth estimation with histogram. The Monocular Depth approach provides fine-grained details but suffers from scale ambiguity, whereas Stereo Matching delivers real-world metrics but with coarser results. We carefully design data generation and distributed matching losses to perform knowledge transfer from monocular to stereo.}
    \label{teaser}
\end{figure}

\section{Introduction}

Stereo matching is one of the fundamental problems in computer vision, which aims to get the disparity of two input images. It is crucial for many downstream problems, such as robotics~\cite{zhang2015building}, autonomous driving~\cite{orb2017slam2,guo2024camera}, and augmented reality~\cite{yang2019security}. With the development of deep learning methods, learning-based methods~\cite{chang2018pyramid,shen2021cfnet,xu2023iterative} have shown dominant performance, but heavily rely on the quality and diversity of the training data. Existing methods mainly use synthetic data and real data for training. However, due to the limitation of the blending engine, most of the synthetic data~\cite{mayer2016large} only contains indoor scenes causing domain gaps between the synthetic data and real-world data, since most of the stereo matching scenarios are outdoor scenes. Besides, the mainstream outdoor stereo datasets~\cite{geiger2012we,menze2015object} are acquired using LiDAR, which is both expensive and time-consuming, sometimes with alignment problems. Because of the certain camera principle and limited view range of the LiDAR camera, the real-world ground-truth labels are usually sparse and incomplete~\cite{guo2024lidar}, causing insufficient supervising signal during fine-tuning limiting the further improvement of the stereo matching model.

Some studies~\cite{yang2019drivingstereo,yao2020blendedmvs, liang2023mpi, guo2024lidar} have tried to boost the generalization and performance of the model by adding more simulated data and real data. Others ~\cite{xu2023iterative,lipson2021raft} improve networks by introducing advanced modules. However, these methods do not solve the most fundamental problems, which are synthetic data domain gap and real-world label sparseness.
To further solve these problems of stereo matching models, we turn our attention to monocular depth estimation, as shown in Figure \ref{teaser}. Recently, lots of researchers have proposed powerful fundamental models in monocular depth estimation like MiDaS~\cite{Ranftl2022}, Megadepth~\cite{MDLi18}, and DepthAnything~\cite{depthanything}, which perform outstanding details and generalization in various datasets. Since their depth estimation output is inverse depth, which has a strong connection with disparities, the great performance of those depth estimation models inspires us to migrate their knowledge to the stereo-matching models to boost their generalization and performance.

To achieve enhanced generalization and performance, we introduce the Mono2Stereo framework, leveraging the strengths of a powerful monocular depth estimation model (MDM). It is utilized during the two-stage training process of stereo matching, comprising synthetic data pre-training and real-world data fine-tuning. Specifically, we first build a stereo training pair generation pipeline to get synthetic training data of real-world scenarios as a pre-training dataset. During generation, we use monocular depth estimation for forward warping for novel view synthesis, then we introduce an Edge-Aware (EA) inpainting module to generate the missing parts in the novel view image. In the second stage, we introduce a Sparse-to-Dense Knowledge Distillation (S2DKD) strategy to further transfer knowledge from the monocular model. Leveraging the powerful relative depth estimation capability of the monocular depth estimation network, our knowledge distillation strategy focuses on supplementing the missing information in sparse labels, such as detail and out-of-view information.

To sum up, we make the following contributions:
\begin{itemize}
\item  We propose a stereo data generation framework that utilizes monocular images with an edge-aware inpainting module, which generates highly realistic stereo data for training deep stereo networks.
\item We propose a Sparse-to-Dense Knowledge Distillation (S2DKD) strategy, which enhances edge detail from monocular depth when fine-tuning on sparse labels.
\item Our method emphasizes the importance of monocular depth for deep stereo networks through extensive experiments. Models trained using our approach achieve state-of-the-art results across various datasets.
\end{itemize}


\section{Related Work}

In this section, we review the most relevant studies on stereo matching, stereo dataset generation, and monocular depth estimation methods.




\subsection*{Stereo Matching Method}

Learning-based stereo matching methods have taken the place of traditional optimization methods with CNN networks. With the development of deep learning, GCNet~\cite{kendall2017end} first uses 3D convolutional encoder-decoder architecture to regularize a 4D volume. Following the success in 3D convolutional network GCNet, PSMNet~\cite{chang2018pyramid}, GwcNet~\cite{guo2019group} and GANet~\cite{zhang2019ga} gradually increment the accuracy of the network. Besides, cascade methods like CFNet~\cite{shen2021cfnet} have been proposed for efficiency improvement. RAFT-Stereo~\cite{lipson2021raft} proposes to recurrently update the disparity field using local cost values retrieved from the all-pairs correlations. IGEV~\cite{xu2023iterative} constructs a new module to encode non-local geometry and context information. In this paper, instead of focusing on network architecture improvement, we focus on training strategies to transfer knowledge from well-trained monocular models. 


\begin{figure*}
    \centering
    \includegraphics[width = \linewidth]{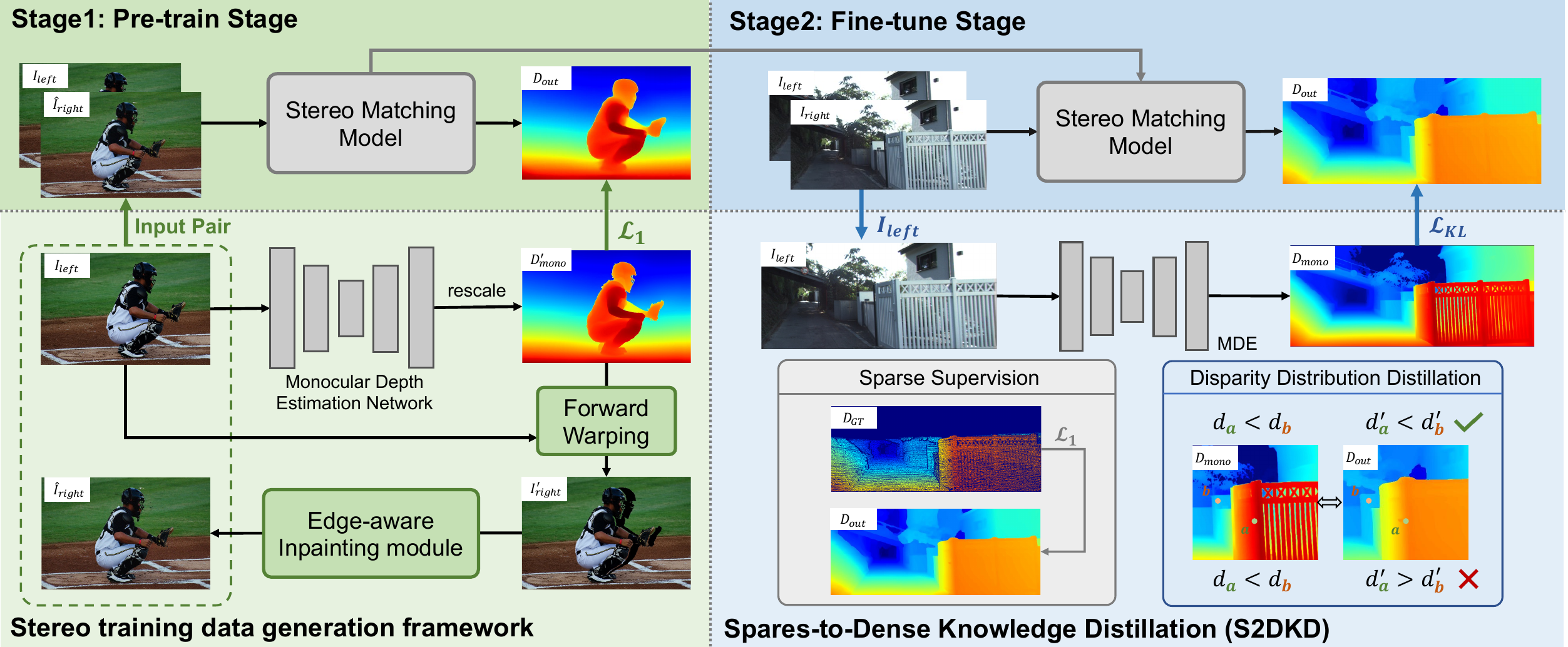}
    \caption{The overall architecture of our proposed method consists of two main stages: (A) Pre-training stage with a data generation pipeline. We estimate and rescale monocular depth maps to construct stereo pairs for training deep stereo networks. (B) Fine-tuning stage with knowledge transfer. We leverage the estimated monocular depth to enhance the prediction of details, which are otherwise lacking in the sparse ground truth.}
    \label{fig:framework}
\end{figure*}

\subsection*{Datasets and Data Generation Method}

Stereo datasets can be roughly divided into synthetic and real-world datasets. For the real-world datasets, KITTI12~\cite{geiger2012we} and KITTI15~\cite{menze2015object} release 200 training pairs of outdoor paired images with sparse disparity labels transformed from LiDAR points. ETH 3D~\cite{schops2017multi} dataset is a representative indoor-scene benchmark with tens of scenes using a depth camera. Recently, there are some large stereo datasets DIML~\cite{cho2021diml}, HRWSI~\cite{xian2020structure}, IRS~\cite{wang2021irs}. However, their GT labels are acquired by stereo matching methods, which limits the upper limit of the performance of deep stereo networks. There are synthetic datasets with the help of computer graphics (CG) to render stereo images like SceneFlow~\cite{mayer2016large}, providing ground truth disparity maps. However, the CG-based method usually fails to model complex real-world scenes, leading to domain gaps with real-world applications. Some work has tried to generate training datasets based on real-world images. Mono-for-Stereo (MfS)~\cite{watson2020learning} generates paired stereo images from single-view images. However, MfS lacks image realism with issues like collisions, holes, and artifacts, which negatively impact the performance of deep stereo networks.  In contrast, our method increment data realism through a well-designed inpainting technique.

\subsection*{Monocular Depth Estimation}

Traditional MDE method mainly rely on handcrafted features and other computer vision techniques, which are limited by diversity and generalization. With the development of deep learning, deep learning based MDE methods have achieved great success. \cite{eigen2014depth} first propose a multi-scale fusion network to predict the depth. Many works have been done to increment networks performance after that. A milestone MiDaS~\cite{Ranftl2022} utilizes an affine-invariant loss to ignore the potentially different depth scales and shifts across varying datasets. Following MiDaS, DepthAnything~\cite{depthanything} further strengthens the model by pay more attention to large-scale monocular unlabeled images and shows strong generalization and refinement in unseen images. In this paper, we aim to increment stereo matching model with the help of DepthAnything.

\section{Method}

In this section, we first introduce the motivation and formulation of our proposed method. Then we present the details of a stereo training data generation framework using monocular images with an edge-aware inpainting module and the Spares-to-Dense Knowledge Distillation (S2DKD) strategy.

\subsection*{Motivation and Formulation}

Our goal is to transfer knowledge from a monocular depth estimation network to a stereo matching network. Specifically, we introduce a two-stage pipeline: (1) pre-training with a data generation framework and (2) fine-tuning on sparse ground truth using a knowledge distillation strategy.

In the first stage, our objective is to generate realistic stereo training data with the help of the monocular model. Given a single-view image $\mathbf{I} \in \mathbb{R}^{H \times W \times 3}$, we assume it to be the left-view image $\mathbf{I}_l \in \mathbb{R}^{H \times W \times 3}$ and generate the corresponding right-view image $\hat{\mathbf{I}}_{r} \in \mathbb{R}^{H \times W \times 3}$ using the ground truth disparity map $\mathbf{D} \in \mathbb{R}^{H \times W}$. This allows us to construct a stereo image training set. The disparity map $\mathbf{D}$ represents the pixel-wise disparity between the left and right view images. However, the disparity $\mathbf{D}_{mono} \in \mathbb{R}^{H \times W} $ obtained from the monocular depth estimation network only contains relative disparity values instead of absolute pixel-wise disparity. We denote this mapped disparity as $\mathbf{D'}_{mono}$. We then propose an edge-aware (EA) inpainting module to handle occlusion holes.

In the second stage, real-world datasets typically contain only sparse ground truth labels, which are insufficient to provide the necessary supervision signals for image details and other critical information. To address this limitation, we introduce a Sparse-to-Dense Knowledge Distillation (S2DKD) strategy to enhance edge accuracy and consistency, particularly in areas concerning fine details and regions outside the view.

\subsection*{Stereo Training Data Generation}


\subsubsection*{Stereo Image Generation with Monocular Depth}

A disparity map is defined as the per-pixel horizontal displacement between the corresponding locations of every pixel from the first view to the second; in this case, from the left image $I_l$ to the right image $I_r$. Described in mathematical form, the disparity map can be defined as
\begin{eqnarray}
\mathbf{D}(i) = x_l(i) - x_r(i')
\end{eqnarray}
where $\mathbf{D}(i)$ is the disparity value of pixel $i$, and $x_l(i)$ and $x_r(i')$ are the horizontal coordinates of pixel $i$ and $i'$ in the left and right view images, respectively.

Given any disparity map, we can warp every left view pixel $i$ to the right view pixel $i'$ using the disparity map. Therefore, we use the monocular depth estimation model to formulate the disparity map $\mathbf{D}_{mono}$. The disparity map from the monocular model is a relative disparity map ranging from $[0,1]$. We scale $\mathbf{D}_{mono}$ with a random factor $f$ to convert it into the pixel-wise disparity map $\mathbf{D}_{mono}'$. The transformation can be formulated as
\begin{eqnarray}
    \mathbf{D}_{mono}' = f \cdot \mathbf{D}_{mono},
\end{eqnarray}
where $f \in [d_{min}, d_{max}]$. To boost the diversity of the generated data, we randomly sample the scaling factor $f$ from a uniform distribution $U(d_{min}, d_{max})$, where $d_{min}$ and $d_{max}$ are the minimum and maximum scaling factors, respectively.


\begin{figure}[t]
\centering
\subfloat[Warping w/o EA]{\includegraphics[width=0.495 \linewidth]{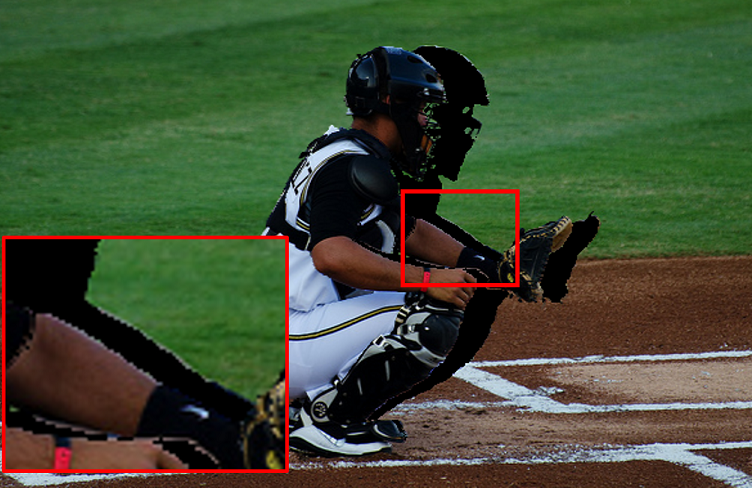}}
\hfill
\subfloat[Inpainting w/o EA]{\includegraphics[width=0.495 \linewidth]{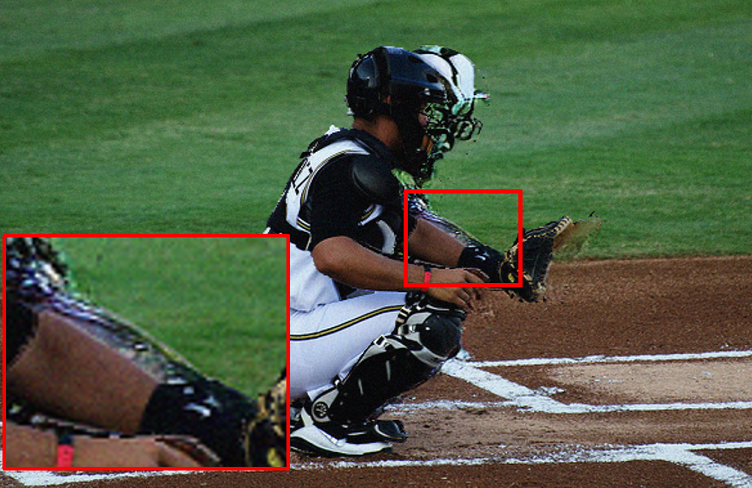}}
\hfill
\subfloat[Warping w/ EA]{\includegraphics[width=0.495 \linewidth]{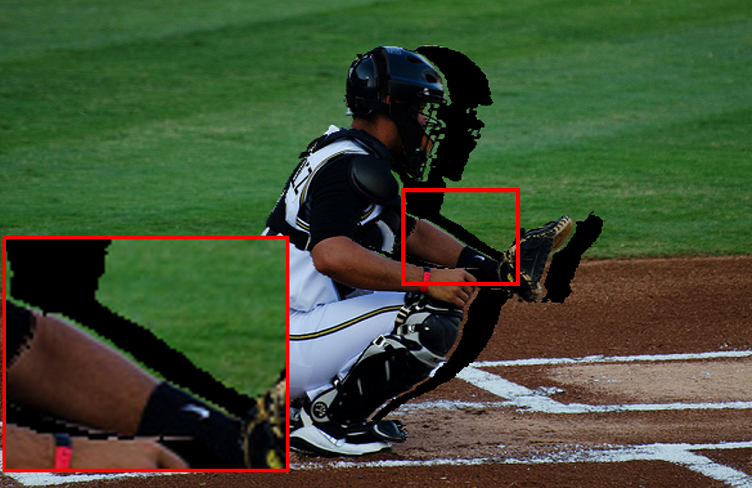}}
\hfill
\subfloat[Inpainting w/ EA]{\includegraphics[width=0.495 \linewidth]{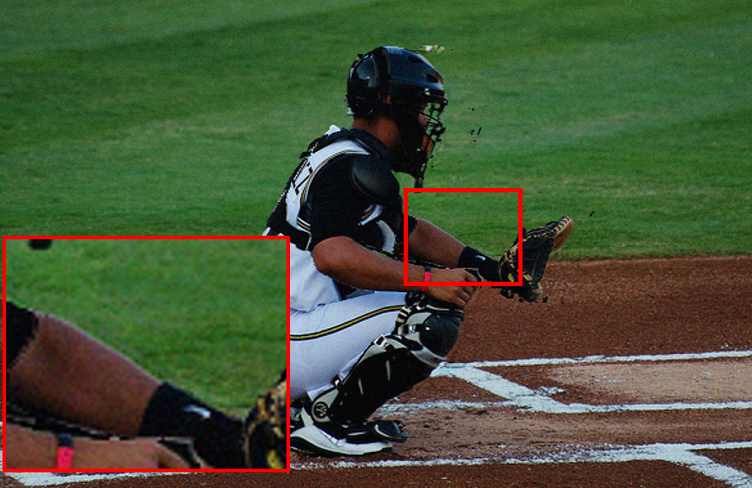}}
\caption{Comparison between naive inpainting module and edge-aware (EA) inpainting module. Our inpainting method generates no artifacts and produces more realistic background images. We present (a) right-view image warping from left-view image with occulusion holes; (b) image inpainted from (a) using Stable Diffusion (SD); (c) right image warping with EA and (d) image inpainted from (c) using SD. }
\label{Inpainting}
\end{figure}

\subsubsection*{Edge-Aware Inpainting Module}

Random scaling and forward warping address the challenge of generating stereo image pairs. However, the generated right-view images often contain occlusion holes in regions that are invisible in the left-view images. To fill in these missing parts, we use Stable Diffusion (SD). However, as shown in Figure \ref{Inpainting}, simply using an inpainting model tends to blend the front and back contents. To address this issue, we propose an Edge-Aware (EA) inpainting module to mitigate the blending problem. According to Figure \ref{Inpainting}, this blending issue arises from a lack of object edge information. Therefore, we first generate the object edge mask $\mathbf{M} \in [0,1]^{H \times W}$ from $\mathbf{D}{mono}'$. We detect the horizontal object edges to create the edge mask using
\begin{eqnarray}
    \mathbf{M}(i,j) = \left\{
    \begin{array}{ll}
    1, & if \  \nabla_x(\mathbf{D}_{mono}') > \tau \\
    0, & otherwise
    \end{array}
    \right.,
\end{eqnarray}
where $\tau$ is the threshold for detecting object edges. After obtaining the mask $\mathbf{M}$, we select a few pixels from the background and warp them with the foreground object to preserve edge information. Next, we perform inpainting using the SD model to generate the final right-view image. Since the edges are preserved, the image inpainted using the EA module significantly reduces the fusion of foreground and background objects, thereby improving the realism of the generated image.

\subsection*{Spare-to-Dense Knowledge Distillation}

Stereo matching models are usually fine-tuned on real-world data after pre-trained on synthetic datasets to get a more robust performance on the target domain. However, the existing outdoor real-world datasets only contain sparse labels due to the limitation of LiDAR, causing details missing during the fine-tuning. To address this issue, we propose a Spares-to-Dense Knowledge Distillation (S2DKD) strategy to provide more detail information on the target domain and more supervising signal during fine-tuning.

\begin{table*}[t]
\centering
\begin{tabular}{@{}llcccccccccc@{}}
\toprule
\multirow{2}{*}{Networks} & \multirow{2}{*}{Training Dataset} & \multicolumn{3}{c}{ETH3D} & \multicolumn{3}{c}{KITTI 12} & \multicolumn{3}{c}{KITTI 15} \\
\cmidrule(lr){3-5} \cmidrule(lr){6-8} \cmidrule(lr){9-11} 
& & EPE $\downarrow$ & D1 $\downarrow$ & \textgreater2px $\downarrow$ & EPE $\downarrow$ & D1 $\downarrow$ & \textgreater2px $\downarrow$ & EPE $\downarrow$ & D1 $\downarrow$ & \textgreater2px $\downarrow$ \\
\midrule
\multirow{3}{*}{PSMNet} & SceneFlow & 8.272 & 8.799 & 10.74 & 4.683 & 30.512 & 43.35 & 5.986 & 32.149 & 44.47  \\
& MFS Dataset & 0.534 & 2.204 & 3.506 & 1.009 & 4.322 & 6.984 & 1.605 & 4.729 & 8.537  \\
&  DiffMFS (Ours) & \textbf{0.397} & \textbf{1.531} & \textbf{2.481} & \textbf{0.859} & \textbf{4.077} & \textbf{6.667} & \textbf{1.052} & \textbf{4.507} &\textbf{8.312}  \\
\midrule
\multirow{3}{*}{CFNet} & SceneFlow  & \textbf{0.413} & 1.832 & 2.629 & 1.010 & 4.742 &  7.789 & 1.839 &  5.758 &9.939 \\
& MFS Dataset & 0.712 & 3.194 & 4.780 & 0.922 & 4.523 & 7.200 & 1.113 & 5.127 & 9.079  \\
&  DiffMFS (Ours) & 0.443 & \textbf{1.434} & \textbf{2.500} & \textbf{0.881} & \textbf{4.169} &  \textbf{6.669} & \textbf{1.030} &  \textbf{4.711} & \textbf{8.562} \\
\midrule
\multirow{3}{*}{IGEV} & SceneFlow & 0.288 & 3.610 & 1.669 & 1.027 & 5.135 & 7.714 & 1.212 & 6.034 & 9.653 \\
& MFS Dataset & 0.703 & 4.657 & 2.105 & 0.824 & 3.546 &5.530 & 1.061 & 4.912 & 8.625 \\
& DiffMFS (Ours) & \textbf{0.284} & \textbf{3.124} & \textbf{1.164} & \textbf{0.803} & \textbf{3.505} & \textbf{5.438} & \textbf{1.001} & \textbf{4.659} & \textbf{8.149}  \\
\bottomrule
\end{tabular}
\caption{Zero-shot validation results of models pre-trained on our generated dataset compared with synthetic dataset and data generation method. The results indicates that our stereo data generation method outperforms the state-of-the-art datasets. Checkpoints of the comparative methods are obtained from OpenStereo \cite{guo2023openstereo} and official implementations.}
\label{tab:zero-shot}
\end{table*}
\begin{figure}[t]
\centering
\subfloat[Ground-Truth Disparity]{\includegraphics[width=0.68 \linewidth]{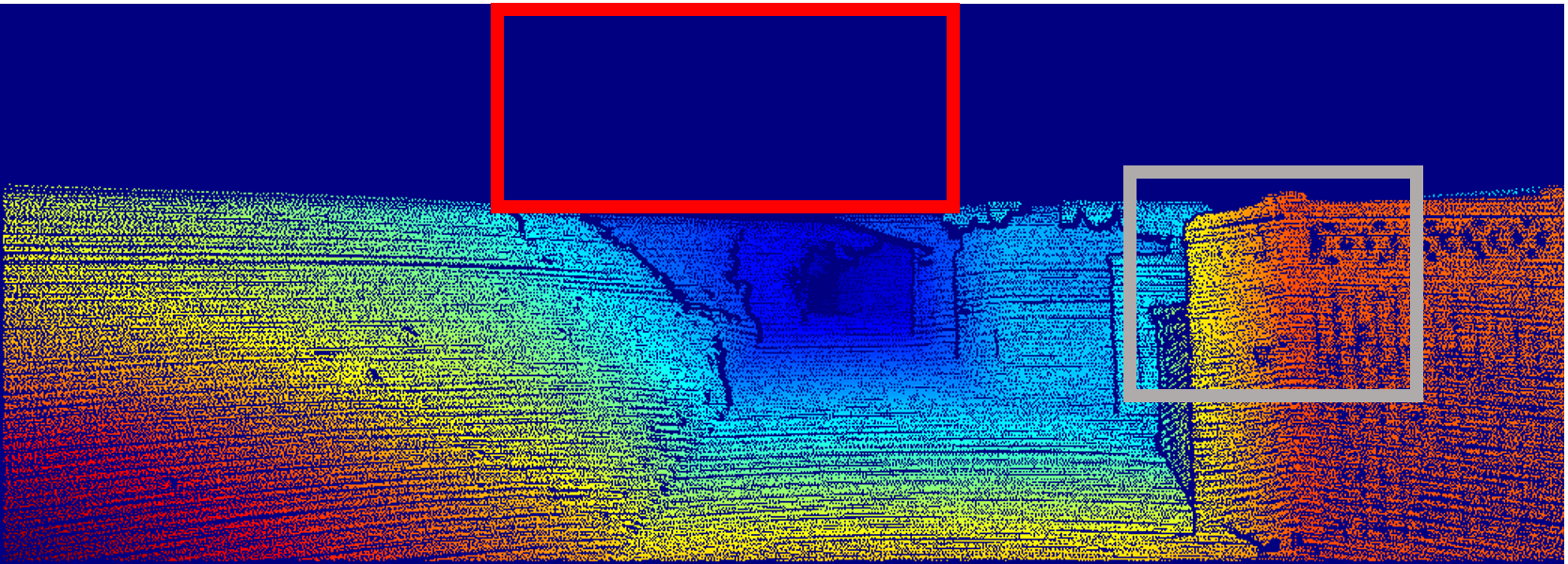}}
\hfill
\subfloat[GT Gradient]{\includegraphics[width=0.31 \linewidth]{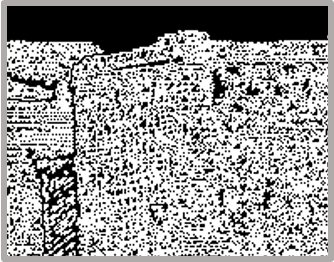}}
\hfill
\subfloat[Monocular Estimation Disparity]{\includegraphics[width=0.68 \linewidth]{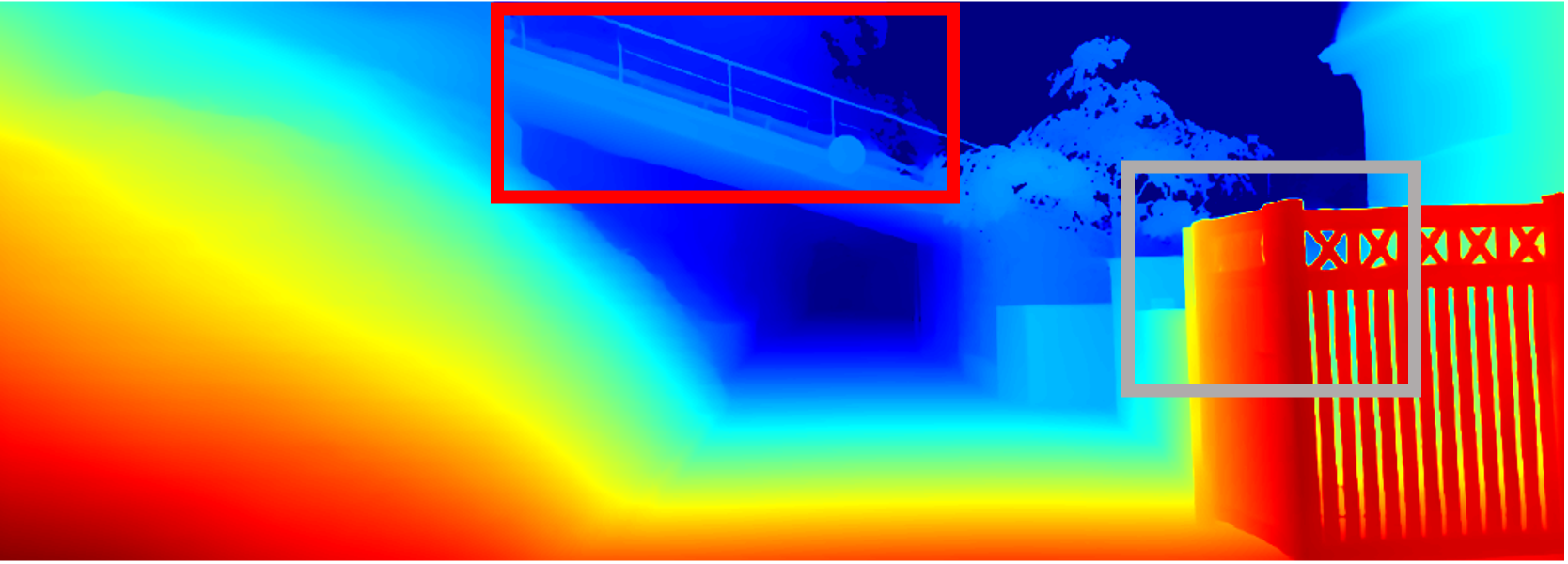}}
\hfill
\subfloat[Mono Gradient]{\includegraphics[width=0.31 \linewidth]{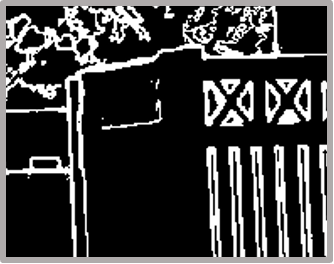}}

\caption{Comparison between disparity from KITTI and from monocular model. We present (a) GT disparity; (b) gradient map calculated using (a); (c) disparity from monocular estimation and (d) gradient map calculated using (c).}
\label{pseudo labels}
\end{figure}
\subsubsection{Disparity Distribution Distillation} As shown in Figure \ref{pseudo labels}, the pseudo-labels generated by monocular depth estimation have better and more accurate details, and can also infer the relative depth of the entire image. However, the depth information generated by monocular depth estimation is not accurate information, but relative depth. Therefore, we cannot directly use the depth map for supervised training.

Therefore, we propose a disparity distribution loss function based on Kullback-Leibler (KL) Divergence. The mathematical form is as 
\begin{equation}
    \mathcal{L}_{KL} =KL(\mathbf{D}_{out},\mathbf{D}_{mono}),
\end{equation}
where $KL(\cdot)$ represents KL divergence. Since the image contains too many pixels,
the value of each pixel after normalization is too small. We found that the gradient disappearance phenomenon often occurs in actual training, which affects the gradient transfer. Therefore, we do not use full-image supervision, but randomly sample pixels for KL supervision. It is formulated as
\begin{equation}
    \mathcal{L}_{KL} =\sum_{i,j}KL(\mathbf{D}_{out}(i,j),\mathbf{D}_{mono}(i,j)).
\end{equation}
Together with the sparse GT loss, the final loss function is formulated as 
\begin{eqnarray}
    \mathcal{L} = \mathcal{L}_{sparse} + \alpha\mathcal{L}_{KL},
\end{eqnarray}
where $\alpha$ is the loss factor that controls the ratio between loss functions.

\section{Experiment}

In this section, we first introduce the datasets we use for training and evaluation, as well as the details of our implementation. Then, detailed comparisons are conducted on supervised and unsupervised settings, respectively. Finally, ablations are conducted to confirm the effectiveness of our proposed main components.

\subsection*{Datasets}

\subsubsection{ETH3D} \cite{schops2017multi} is another commonly used dataset with grayscale stereo pairs from indoor and outdoor environments, again with LiDAR ground truth. ETH3D contains 27 training and 20 test frames (data/results) for low-resolution two-view stereo on frames of the multi-camera rig. We take ETH3D as another validation set to verify the effectiveness of our proposed method.

\subsubsection{KITTI 2012 and KITTI 2015} KITTI 2012 \cite{geiger2012we} and KITTI 2015 \cite{menze2015object} are renowned benchmarks for stereo matching, each with 200 labeled pairs for training and additional pairs for testing. The ground truth is given by sparse LiDAR. Compared to the KITTI 2012 and KITTI 2012 benchmarks, KITTI 2015 comprises dynamic scenes for which the ground truth has been established in a semi-automatic process. By default, we evaluate the trained models on the training sets of KITTI 2012 and KITTI 2015, which serve as the validation sets, following previous work. Additionally, we present the fine-tuned results on the test sets of the KITTI 2015 official benchmark.

\subsubsection{Training datasets} Training datasets can be divided into two categories. One is synthetic data, represented by SceneFlow \cite{mayer2016large}, on which most supervised deep stereo networks are trained. The ScaneFlow dataset contains more than $39,000$ stereo frames in $960\times540$ pixel resolution, rendered from various synthetic sequences. However, this inevitably creates domain gaps from real-world applications. Therefore, we turn to train the models on a combination of varied single image datasets, which is called the Diffusion-based Mono for Stereo (\textbf{DiffMFS}) dataset. Following \cite{watson2020learning}, we collect all the training images from COCO 2017 \cite{lin2014microsoft}, Mapillary Vistas \cite{neuhold2017mapillary}, ADE20K \cite{zhou2017scene}, Depth in the Wild \cite{chen2016single}, and DIODE \cite{vasiljevic2019diode}. This results in $597,727$ in monocular images. We then use our proposed method to synthesize stereo data for training stereo matching models on these images.

\subsection*{Implementation Details}

\noindent\textbf{Networks Architectures.} To evaluate how effective our generated data are at training deep stereo networks, we take the widely-used PSMNet variant \cite{chang2018pyramid}, which represents one of the state-of-the-art architectures for supervised stereo matching and has excellent generalization capability. We also show the results of another two representative network architectures, including CFNet \cite{shen2021cfnet} and IGEV \cite{xu2023iterative}. For monocular depth estimation, we use Depth-Anything \cite{depthanything}, which represent the state-of-the-art. 

\noindent\textbf{Evaluation Metrics.} For evaluation metrics, we report evaluation results on the average End-Point Error (EPE). Besides, we also report the D1 and $>$2px metric, which indicates the percentage of stereo disparity outliers in the first frame. For both metrics, smaller values indicate better model performance.


\subsection*{Comparison with Data Generation Methods}
As shown in Table \ref{tab:zero-shot}, we validate the performance of our proposed stereo data generation framework on public benchmarks, comparing it with another dataset and data generation method. We start with a comparison with the synthetic dataset SceneFlow \cite{mayer2016large}, which represents the most advanced data synthesis methods. Besides, as a dataset generation method from real-world images, our generated dataset DiffMFS is compared with MFS \cite{watson2020learning}, which is the state-of-the-art real-world dataset generation technique. For fair comparisons, we conduct experiments with the same setup as our competitor. 

We conduct zero-shot experiments where three types of model architectures are trained on one generated dataset. As shown in Table \ref{tab:zero-shot}, with the same setup of generated data, our method achieves significant improvements and better generalization across multiple datasets and network architectures. It is worth noting that our approach is particularly prominent on outdoor KITTI datasets. The experimental results show the superiority of the proposed data synthesis method. Since IGEV shows the best performance, we adopt it as our default network structure.

\begin{figure*}[t]
    \centering
    \subfloat[Left Frames]{\includegraphics[width=0.24 \linewidth]{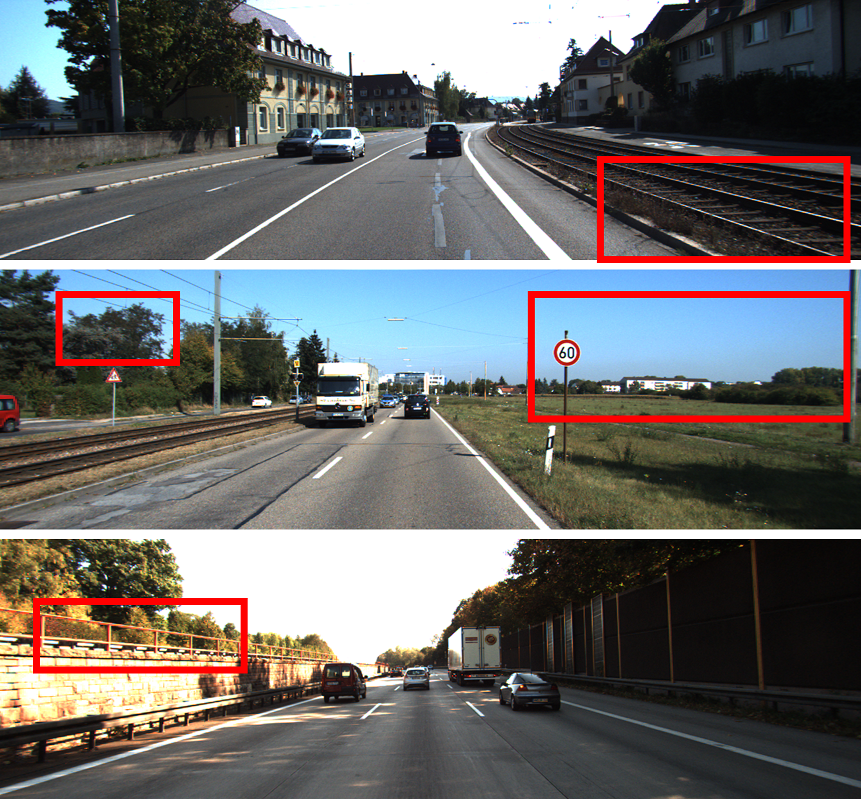}}
    \hfill
    \subfloat[IGEV]{\includegraphics[width=0.24 \linewidth]{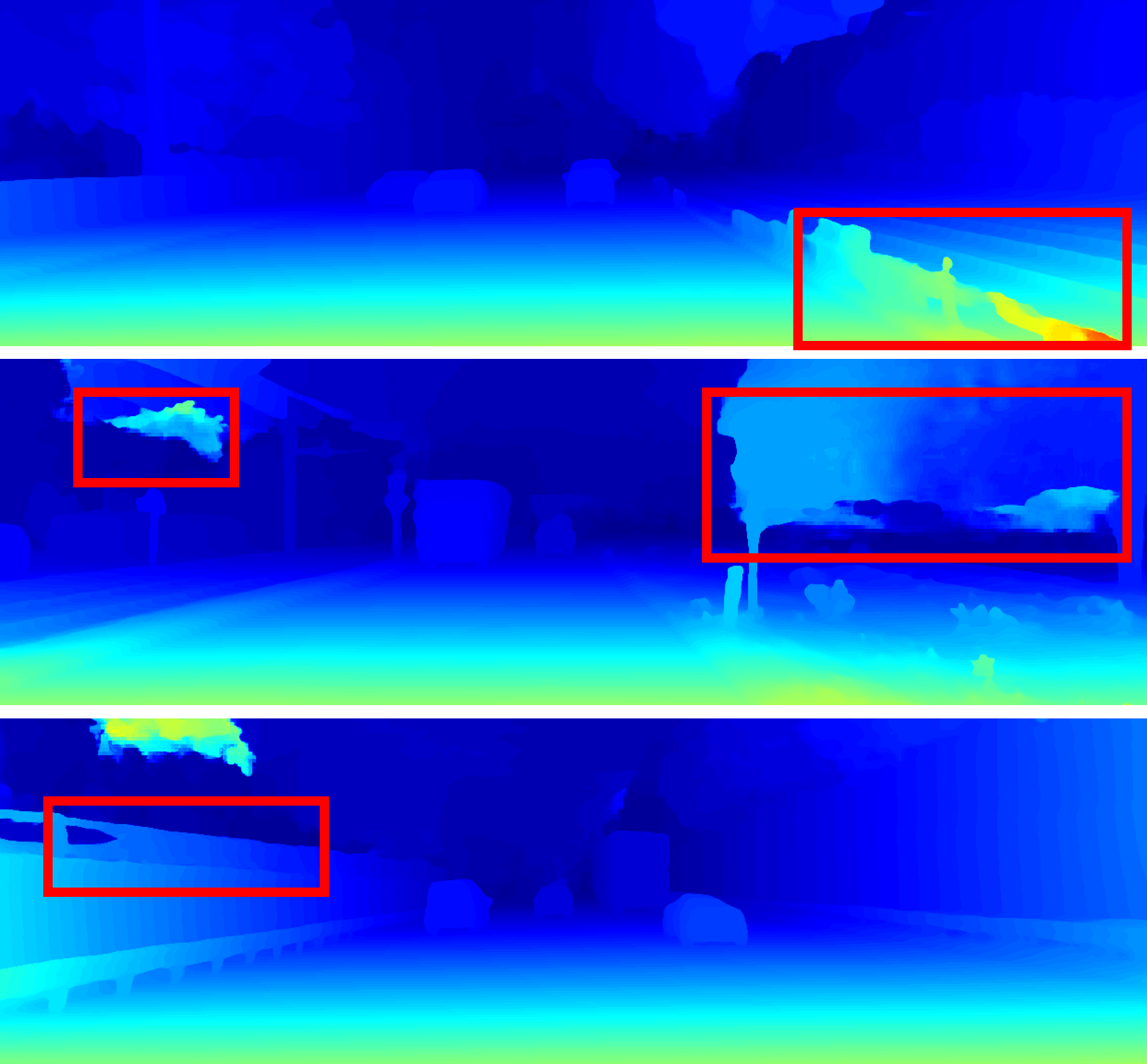}}
    \hfill
    \subfloat[IGEV + DiffMFS]{\includegraphics[width=0.24 \linewidth]{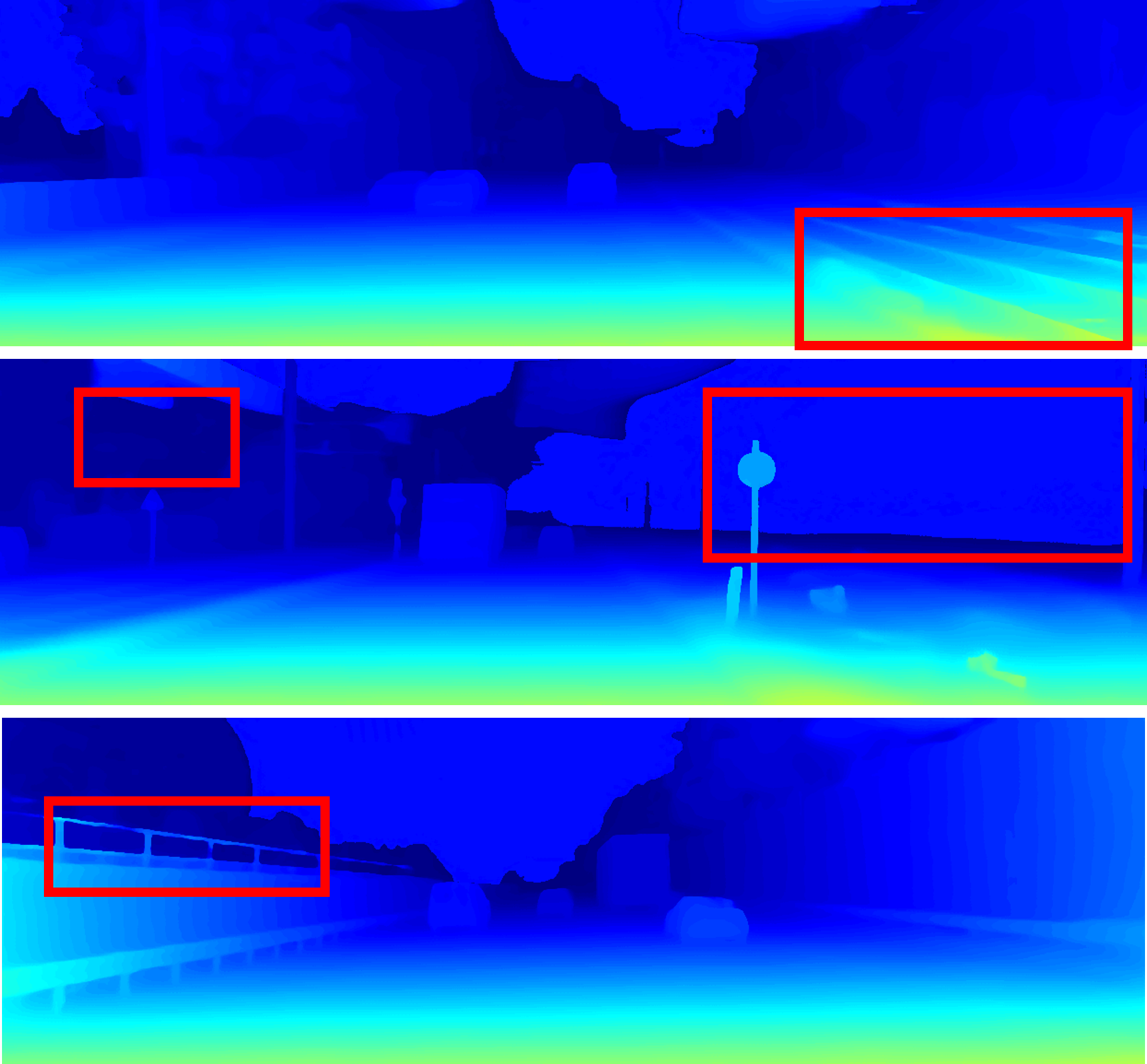}}
    \hfill
    \subfloat[IGEV + DiffMFS + S2DKD]{\includegraphics[width=0.24 \linewidth]{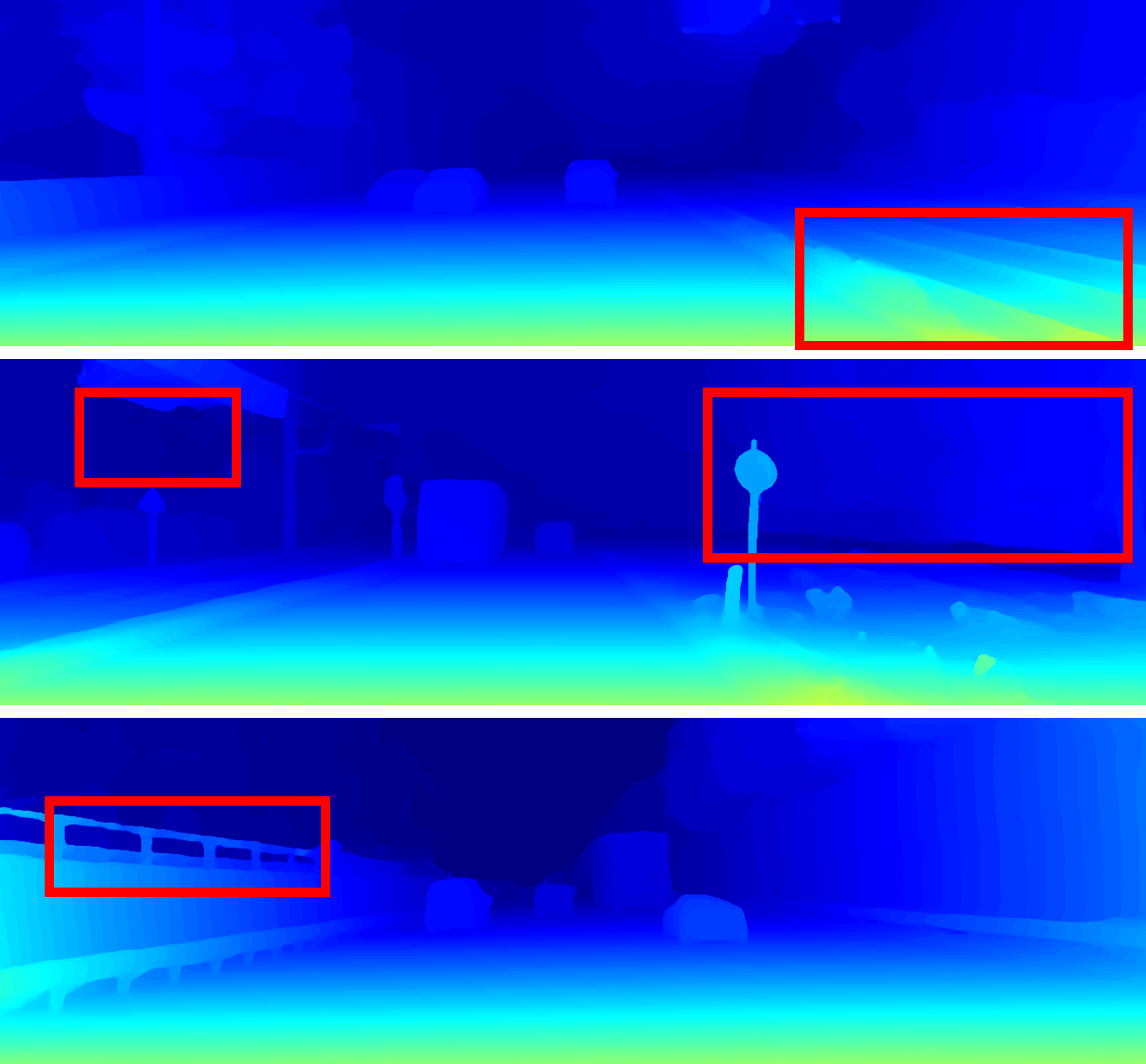}}
    \caption{Qualitative results of IGEV trained with and without our proposed S2DKD strategy and our generated DiffMFS dataset. The default model is pre-traiend on sceneflow and fine-tuned on the KITTI training set by official implementation.}
    \label{fig:vis-dataset}
\end{figure*}

\begin{figure}[t]
    \centering
    \includegraphics[width = \linewidth]{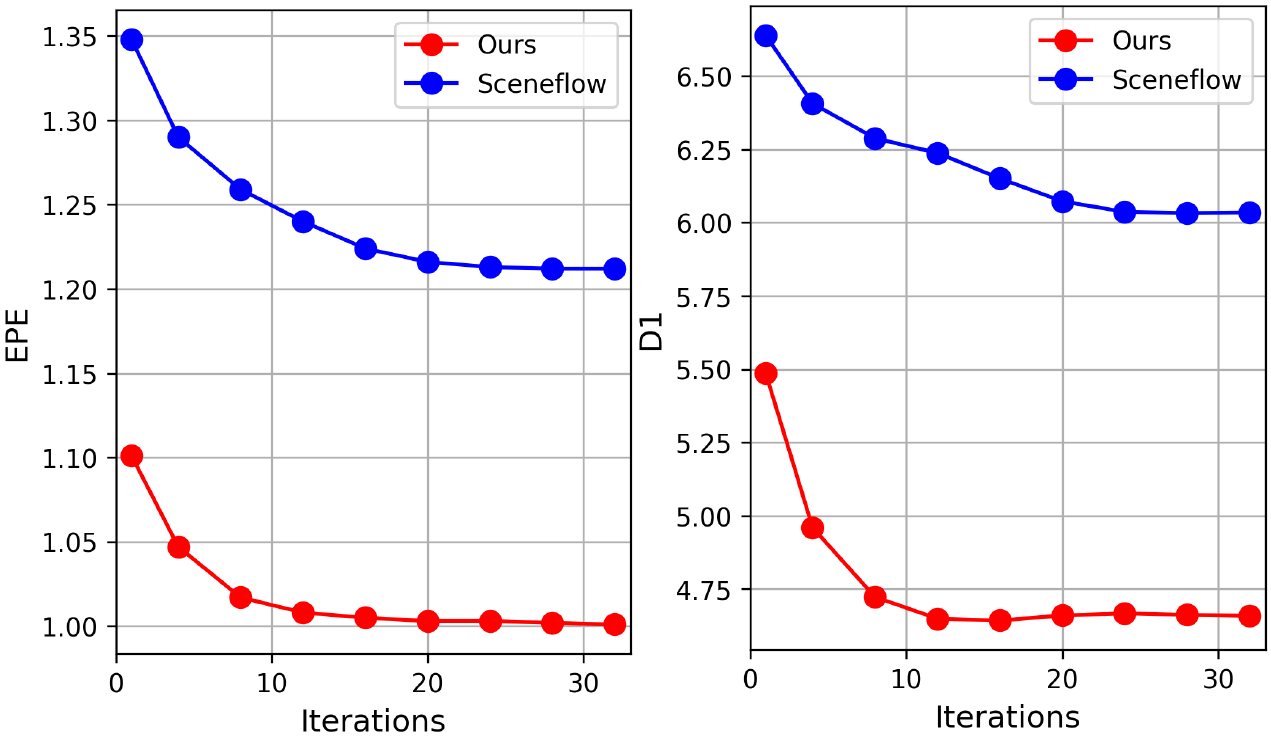}
    \caption{Performance on KITTI 15 with different number of refinements at inference time. Our method brings earlier stability and therefore fewer refinements.}
    \label{fig:line}
\end{figure}

\begin{table}[t]
\centering
\begin{tabular}{@{}lcc@{}}
\toprule
\multirow{2}{*}{Methods} & \multicolumn{2}{c}{KITTI 15} \\
\cmidrule(lr){2-3} 
& EPE $\downarrow$ & D1 $\downarrow$ \\
\midrule
MonoDepth & 1.96 & 10.86 \\
SeqStereo & 1.84 & 8.79 \\
StereoUnsupFt→Mono & 1.71 & 7.06 \\
BridgeDepthFlow & 1.57 & 6.01 \\
Flow2Stereo & 1.34 & 6.13 \\
FLC & 1.22 & - \\
UFD-PRiME & 1.99 & - \\
NeRF-Stereo & 1.44 & 4.83 \\
Ours & \textbf{1.05} & \textbf{4.50} \\
\bottomrule
\end{tabular}
\caption{Comparison with unsupervised methods, in which labels from the KITTI datasets are not used. By default, all unsupervised methods have undergone training process with only images from the KITTI datasets available. PSMNet is used as most previous researches do.}
\label{tab:unsupervised}
\end{table}

\subsection*{Comparison with Unsupervised Methods}

Table \ref{tab:unsupervised} presents the results of a comparative analysis of various unsupervised learning methods on the KITTI 15 test set, including \cite{godard2017unsupervised, yang2018segstereo, guo2018learning, lai2019bridging, liu2020flow2stereo, chi2021feature, yuan2023ufd, tosi2023nerf}. In this experiment, all the unsupervised methods are trained without access to ground truth labels from the KITTI 15 training set, relying solely on the images. This setup ensures a fair comparison across different methods, as none of the models had access to ground truth flow data. Our model follows the same training protocol and is trained on our FA-Flow Dataset. As shown in the table, it outperforms other methods by a notable margin. Specifically, our model achieves an EPE score of $1.05$ and a D1 score of $4.50$, which is much lower than the closest competitor. These results also demonstrate the superior generalization of our model to out-of-domain images, highlighting its effectiveness in unsupervised stereo matching.

\subsection*{Qualitative and Quantitative Result}

Figure \ref{fig:vis-dataset} illustrates the qualitative results of stereo predictions from models trained on different datasets. In the first row, the left frame shows a scene with a road, building, and railing. The IGEV depth map suffers from artifacts and inaccuracies, particularly near the railing and road edges. With the addition of DiffMFS, the depth estimation improves, reducing artifacts and providing clearer object boundaries along the railing. When S2DKD is added, further improvements are seen in depth accuracy and clarity, with cleaner and more detailed depth maps. In the second and third rows, similar trends are observed—IGEV initially shows noticeable artifacts around complex geometries like poles and signs, but DiffMFS and S2DKD incrementally enhance the depth accuracy and object delineation, resulting in clearer and more refined depth maps with fewer artifacts.

The graph in Figure \ref{fig:line} illustrates the advantages of our method compared to Sceneflow on the KITTI 15 dataset during inference. Specifically, our method demonstrates earlier stability in both EPE and D1 metrics, resulting in fewer refinement iterations needed to reach convergence. This leads to faster performance with lower overall error across iterations, as shown by the rapid decline and stabilization of the red curves. In contrast, Sceneflow requires more iterations to stabilize and performs less efficiently, maintaining higher error values throughout the process. This efficiency in reaching stability highlights the robustness and computational advantage of your approach.



\begin{table}[t]
\centering
\begin{tabular}{@{}llcc@{}}
\toprule
\multirow{2}{*}{Experiments} & \multirow{2}{*}{Methods} & \multicolumn{2}{c}{KITTI 15} \\
\cmidrule(lr){3-4}
& & EPE $\downarrow$ & D1 $\downarrow$ \\
\midrule
\multirow{4}{*}{S2DKD} & Off & 0.607 & 1.521 \\
& Grad & 0.626 & 1.564 \\
& L2 & 0.601 & 1.528  \\
& \underline{KL} & \textbf{0.600} & \textbf{1.459}  \\
\midrule
\multirow{3}{*}{Inpainting} & Random & 1.049 & 4.832\\
& SD & 1.032 & 4.738 \\
& \underline{SD + EA} & \textbf{0.992} & \textbf{4.529}\\
\midrule
\multirow{2}{*}{MDE} & MiDaS &  1.023 & 4.689\\
& \underline{Depth-Anything} &  \textbf{0.992} & \textbf{4.529}\\
\bottomrule
\end{tabular}
\caption{Ablation experiments. Settings used in our final framework are \underline{underlined}. $20\%$ data of the KITTI 15 is used as the validation set, the rest of the data and all of the KITTI 12 are used for training, due to the limitation of number of submissions on the official benchmark. ``EA" indicates our proposed edge-aware inpainting module. Fine-tuning is not performed when conducting experiments on inpainting models and monocular depth models.}
\label{tab:ablations}
\end{table}

\subsection*{Ablations}

In this section, we conduct a series of ablation studies to analyze the impact of different module choices in our method. We measure all the factors using $20\%$ data of the KITTI 15 for validation. The rest data of the KITTI 15 and all of the KITTI 12 are used for training, due to the limitation of the number of submissions on the official benchmark.

\subsubsection{Loss in S2DKD.} When training on data with sparse labels, we propose the S2DKD strategy to match the dense monocular depth and the prediction results from the stereo networks. Considering the inconsistency between monocular depth and stereo prediction distribution with metric, we can use multiple losses for distribution matching. As in Table \ref{tab:ablations}, a remarkable improvement is achieved by using either the KL loss, compared with direct fine-tuning without using our proposed S2DKD strategy. Direct use of L2 or gradient losses can have adverse effects due to the mismatched scale between monocular depth and stereo matching. In contrast, KL is more concerned that distribution is not sensitive to size, leading to better performance.

\subsubsection{Inpainting} When using our data generation method, the hole in the newly generated image means that there is no pixel matched to the monocular image. We use the Stable-Diffusion Inpainting model for inpainting these holes. As summarized in Table \ref{tab:ablations}, using an advanced inpainting model leads to better performance when compared to the random hole-filling strategy used in \cite{watson2020learning}. However, Stable Diffusion tends to create confusion between the object and the background edge. We solve this problem by the proposed Edge-Aware filling module and achieve better performance, leading to more visually reasonable content.

\subsubsection{Monocular Depth} We compare monocular depths from two models, including MiDaS \cite{ranftl2020towards} and Depth-Anything \cite{depthanything}. These two models are both robust monocular depth estimation models trained on large-scale monocular datasets. Compared to MiDaS, Depth-Anything achieves better performance through self-monitoring. The experimental results in Table \ref{tab:ablations} also show that the use of a more powerful monocular model brings slightly better performance to the stereo networks.

\section{Conclusion}

In this work, we address the challenges posed by the domain gap between synthetic data and real-world stereo data with sparse labels in deep stereo networks. By introducing \textbf{Mono2Stereo}, a novel approach that leverages monocular depth knowledge to enhance stereo matching networks. Our two-stage training process, which includes synthetic data pre-training with a stereo training data generation pipeline and real-world data fine-tuning with Sparse-to-Dense Knowledge Distillation (S2DKD), has shown significant improvements in both zero-shot generalization and in-domain performance. Through the use of a monocular depth estimation model, our approach allows for a more detailed understanding of scene geometry. Additionally, the forward warping technique used in novel view synthesis and the Edge-Aware inpainting module help bridge gaps between synthetic and real-world scenario, making the model more resilient to challenging scenarios. The experimental results validate the effectiveness of our approach, particularly in overcoming the limitations of sparse real-world labels and enhancing the overall consistency of stereo predictions. Furthermore, the results highlight the potential for broader applications, such as autonomous driving and augmented reality, where stereo vision systems can benefit from the seamless integration of monocular depth estimation and stereo matching networks.


\bibliography{aaai25}

\end{document}


\maketitle

%

\section{A. Experiments on the KITTI15 Testset}

In this section we provide quantitative and qualitative indicators results on the KITTI testset.

\begin{table}[t]
\centering
\begin{tabular}{@{}lccc@{}}
\toprule
\multirow{2}{*}{Methods} & \multicolumn{3}{c}{KITTI 15 (test)} \\
\cmidrule(lr){2-4} 
& D1-bg $\downarrow$ & D1-fg $\downarrow$ & D1-all $\downarrow$ \\
\midrule
PSMNet & 1.86 & 4.62 & 2.32 \\
GwcNet & 1.74 & 3.93 & 2.11 \\
GANet-deep & 1.48 & 3.46 & 1.81 \\
AcfNet & 1.51 & 3.80 & 1.89 \\
HITNet & 1.74 & 3.20 & 1.98 \\
EdgeStereo-V2 & 1.84 & 3.30 & 2.08 \\
LEAStereo & 1.40 & 2.91 & 1.65 \\
ACVNet & \textbf{1.37} & 3.07 & 1.65  \\
CREStereo & 1.45 & 2.86 & 1.69 \\
RAFT-Stereo & 1.58 & 3.05 & 1.82 \\
IGEV & 1.38 & \textbf{2.67} & 1.59 \\
IGEV + Ours & \textbf{1.36} & \textbf{2.67} & \textbf{1.58} \\
\bottomrule
\end{tabular}
\caption{Comparison with supervised methods. ``+ Ours" indicates pre-training on our DiffMFS dataset and fine-tuning with our proposed sparse-to-dense knowledge distillation strategy. Models are validated on the KITTI 15 test sets using official benchmark. By default, all supervised methods have undergone SceneFlow pre-training process and then fine-tune on training sets of KITTI 12 and KITTI 15.}
\label{tab:finetune-test}
\end{table}

\subsection{A1. Comparison with Supervised Methods}

To validate the effectiveness of our proposed Sparse-to-Dense Knowledge Distillation strategy and the generalization capability of our model, we compare it with several supervised methods~\cite{chang2018pyramid,guo2019group,zhang2019ga,zhang2019acfnet,tankovich2021hitnet,song2020edgestereo,cheng2020hierarchical,xu2022attention,li2022practical,lipson2021raft}, as shown in Table \ref{tab:finetune-test}. Following the Settings of the previous work, the supervised models are first pre-trained on the SceneFlow dataset and then fine-tuned on the training sets of KITTI 12 and KITTI 15. This training combination indicates the state-of-the-art training process to date. The model is first pre-trained by synthesizing stereo data from monocular images. Then, the model is fine-tuned in the training sets of KITTI 12 and KITTI 15. Given that KITTI contains only sparse truth values, the sparse-to-dense knowledge distillation strategy proposed by us is added to improve the margin and consistency of prediction results when fine-tuning. As shown in Table \ref{tab:finetune-test}, our fine-tuning method has reached the level of state-of-the-art methods. 

In addition, we propose to use the generated disparity map to warp the left image to obtain the right image. According to the definition of disparity, we measured the PSNR between the newly generated right image and the real right image. The metrics of our method are significantly higher than the simple fine-tuning method.

\begin{table}[t]
\centering
\begin{tabular}{@{}lcc@{}}
\toprule
\multirow{2}{*}{Methods} & \multicolumn{2}{c}{KITTI 15 (test)} \\
\cmidrule(lr){2-3} 
& PSNR $\downarrow$ & SSIM $\uparrow$ \\
\midrule
IGEV & 12.166 & 0.306\\
IGEV$^*$& 12.181 & 0.306\\
IGEV + Ours & \textbf{12.338} & \textbf{0.311}\\
\bottomrule
\end{tabular}
\caption{Quantitative metrics of PSNR of generated right images using predicted disparity and ground truth right images.}
\label{psnr:finetune-test}
\end{table}

\subsection{A2. Qualitative results on KITTI15 Testset}

\begin{figure*}[t]
\centering
\subfloat[Left-view Image]{\includegraphics[width=0.32 \linewidth]{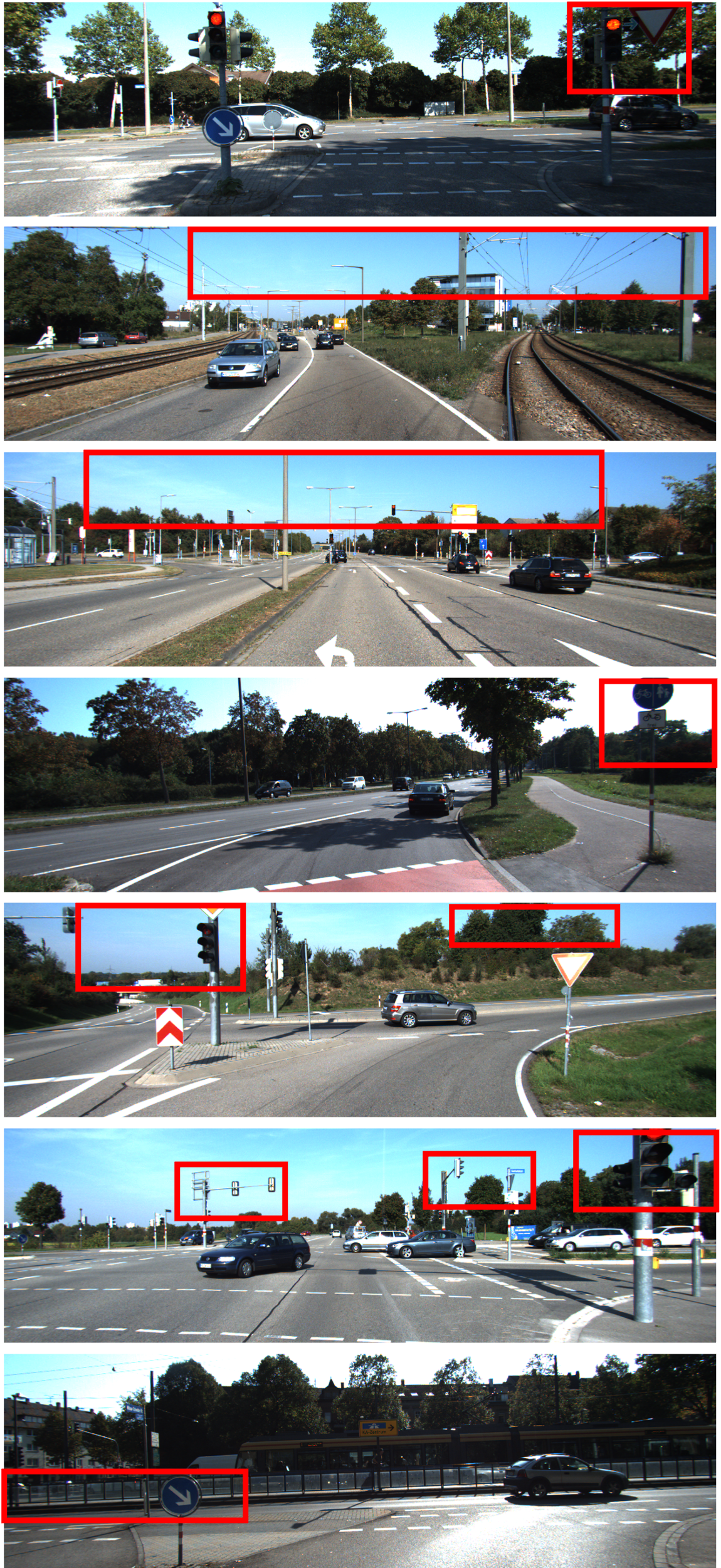}}
\hfill
\subfloat[IGEV]{\includegraphics[width=0.32 \linewidth]{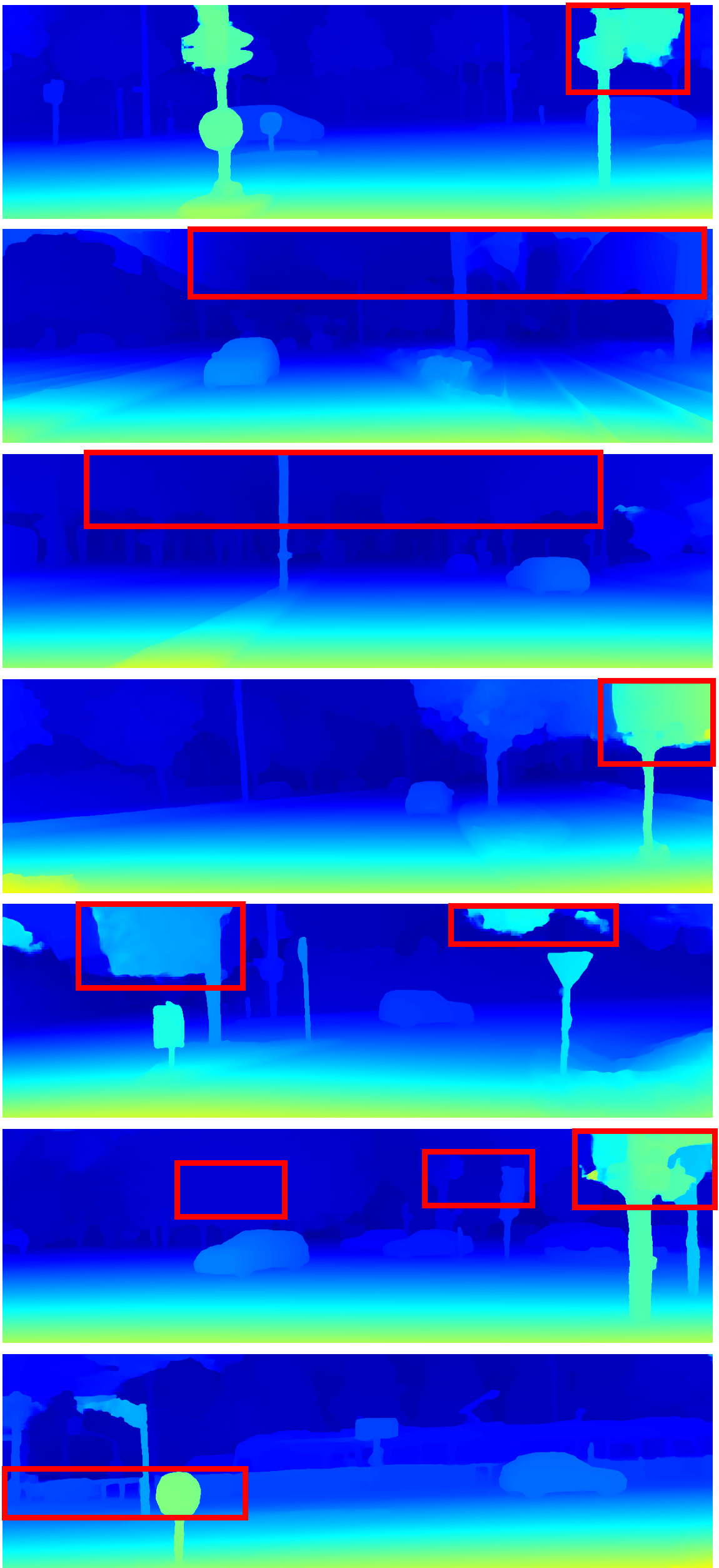}}
\hfill
\subfloat[IGEV+Ours]{\includegraphics[width=0.32 \linewidth]{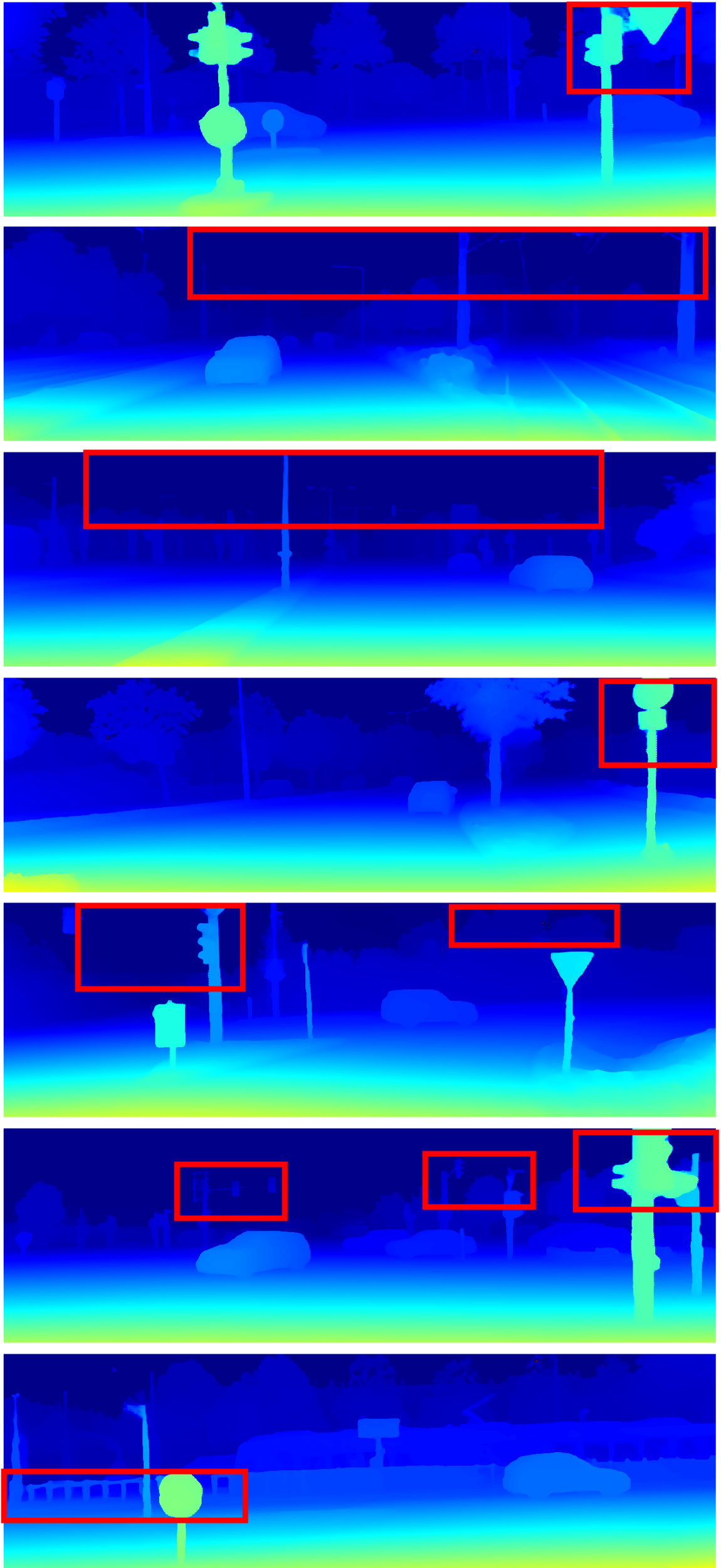}}
\caption{Qualitative results of IGEV on KITTI testset trained with and without our proposed S2DKD strategy and our generated DiffMFS dataset. The default model is pre-traiend on sceneflow and fine-tuned on the KITTI training set by official implementation.}
\label{vis-testset}
\end{figure*}

We show the visualization results of different methods on the KITTI test set in Figure \ref{vis-testset}. Our main advantage is reflected in places where KITTI labels do not exist, such as the sky and distant objects. The poor performance of other methods on the sky makes us question whether those methods have overfitted in the following areas during the fine-tuning process. However, the monocular deep network pseudo-labels we introduced are valid labels for the entire image, so they can avoid predicting unrealistic disparity values in these areas. Unfortunately, these advantages cannot be reflected on the KITTI testset benchmark because these areas are unknown to the KITTI dataset, which explains why our method does not achieve significant improvement on the benchmark.

\section{B. Details about Edge-Aware Inpainting Module}

\begin{figure*}[t]
    \centering
    \includegraphics[width=\linewidth]{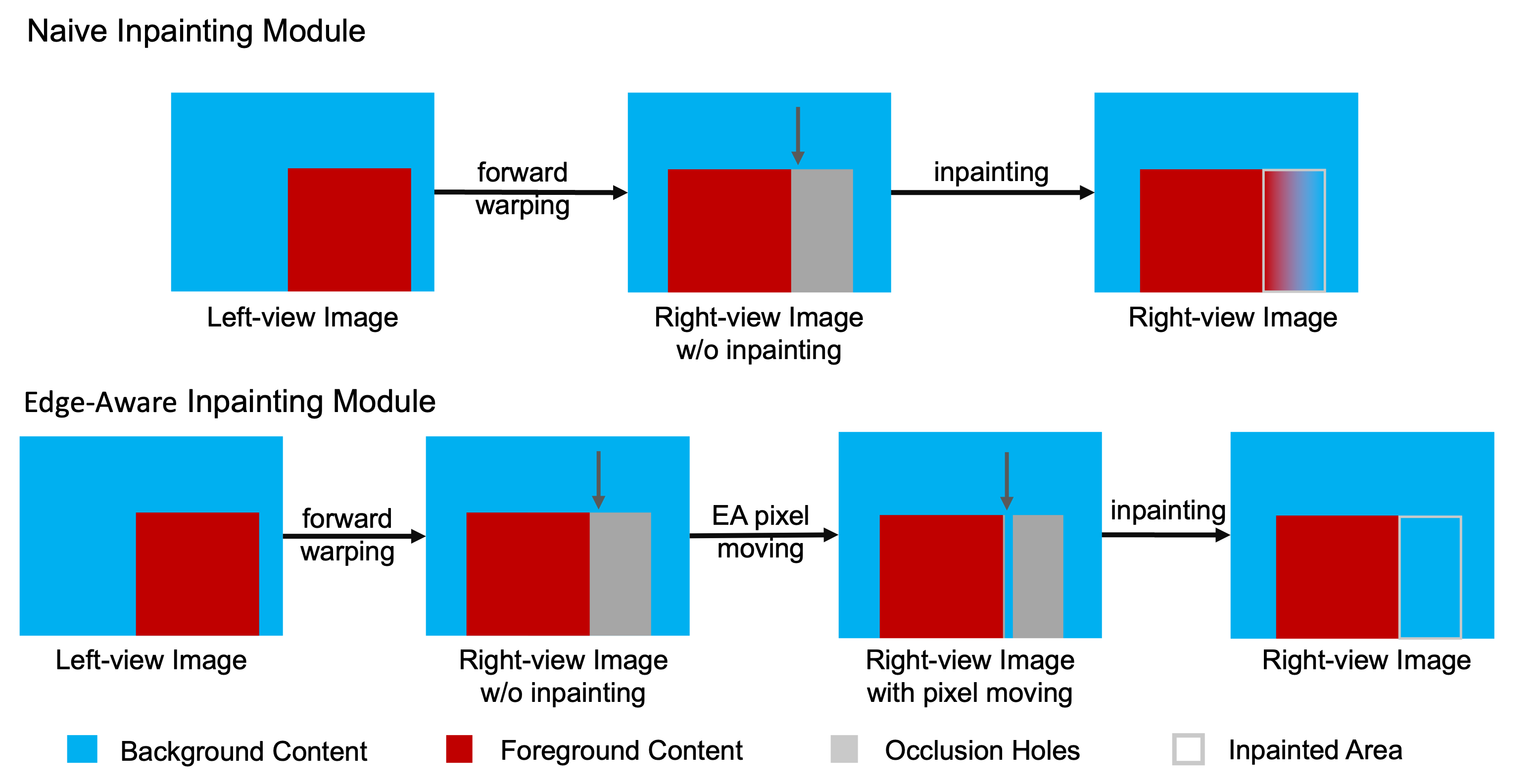}
    \caption{Schematic diagram of the algorithm flow of different inpainting methods. This figure shows that our inpainting method can generate more reasonable completion information. The \textcolor{blue}{blue part} and \textcolor{red}{red part} represents the \textcolor{blue}{background} and \textcolor{red}{foreground} object of the picture, and the \textcolor{gray}{gray part} represents the content that needs to be completed after warping and the inpainted outcome. The content pointed by the arrow represents the area of our EA pixel moving, which is 2 pixels in our implementation.}
    \label{Inpainting}
\end{figure*}

In this section, we further analyze the principle of the proposed inpainting module and give more visualization results to prove the effectiveness of the inpainting module.

\subsection{B1. Discussion on the Method Effectiveness}

The reason why we propose the inpainting module is that the content generated by directly using the inpainting stable diffusion module for completion contains both foreground and background, resulting in mixed content textures. According to Figure \ref{Inpainting}, we analyzed that this phenomenon occurs because the inpainting model cannot effectively judge the boundaries after warping, resulting in a tendency to generate mixed content. Therefore, we effectively hint the inpainting model by preserving some boundaries, thereby avoiding the generation of completion content containing mixed textures.

\subsection{B2. Qualitative of Edge-Aware Inpainting Module}

\begin{figure*}[ht]
\centering
\subfloat[Warped Image]{\includegraphics[width=0.32 \linewidth]{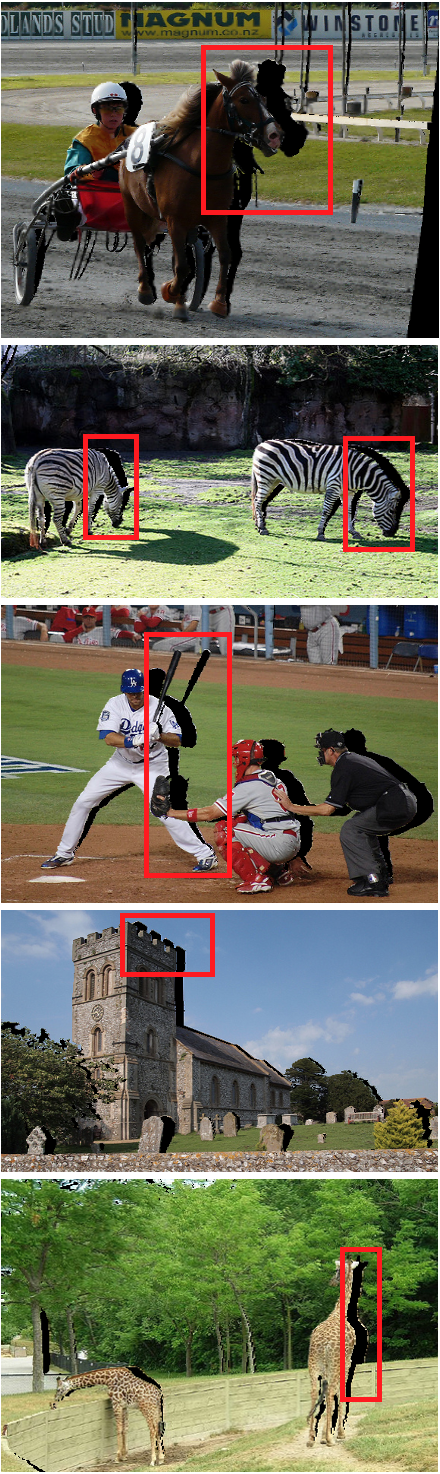}}
\hfill
\subfloat[Inpainting w/o EA]{\includegraphics[width=0.32 \linewidth]{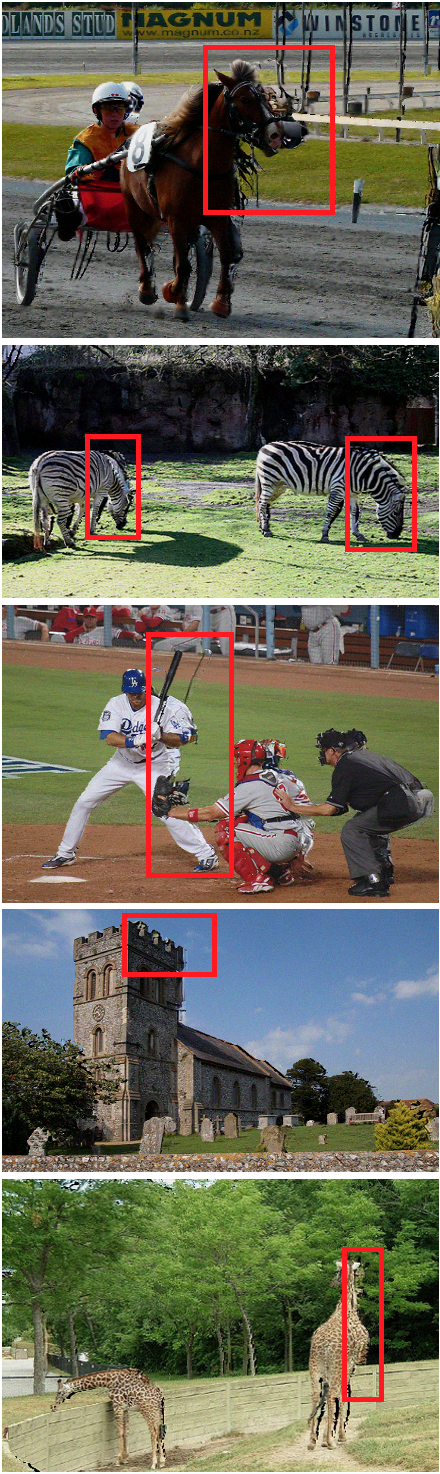}}
\hfill
\subfloat[Inpainting w/ EA]{\includegraphics[width=0.32 \linewidth]{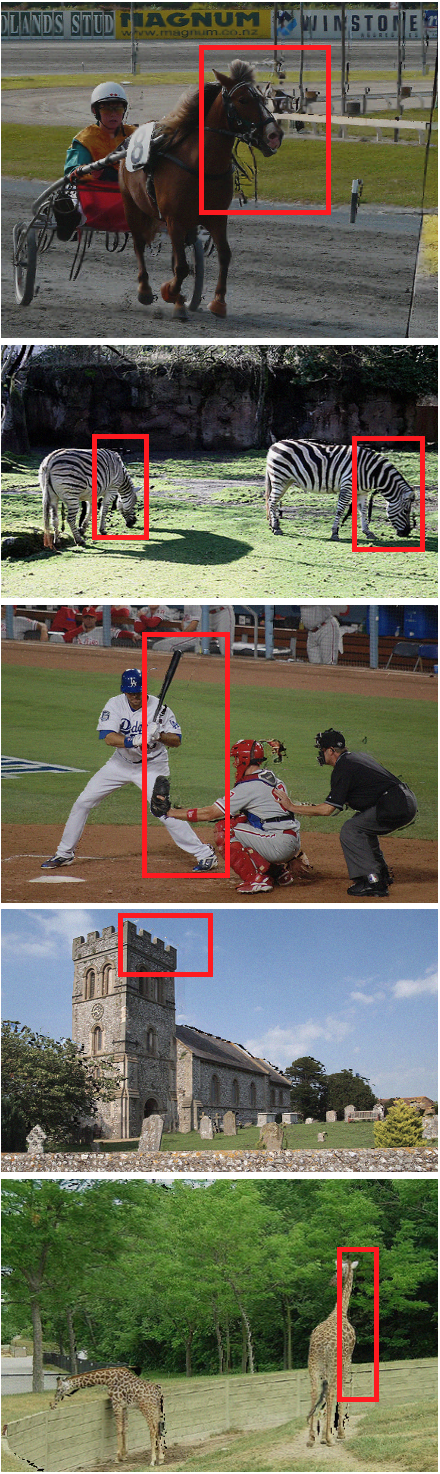}}
\hfill

\caption{Visulization results of inpainted right-view images with or without our method. The results without our method have obvious artifacts and strange textures, while the results of inpainting with our method are more realistic.}
\label{Inpainting-vis}
\end{figure*}

Figure \ref{Inpainting-vis} focuses on the visual comparison between our inpainting method and the direct inpainting method. The direct inpainting method cannot effectively distinguish the foreground and background information of the image, resulting in extremely unrealistic artifacts and strange textures. However, our method can achieve better background completion and avoid the generation of the above unrealistic completion content.

\section{C. Qualitative Results}

In this section we show more visualizations of the pretrain method and the fine-tuned model. The training results show the superiority of our method.

\subsection{C1. Zero-shot Qualitative Results}

\begin{figure*}[ht]
\centering
\subfloat[Left-view Image]{\includegraphics[width=0.32 \linewidth]{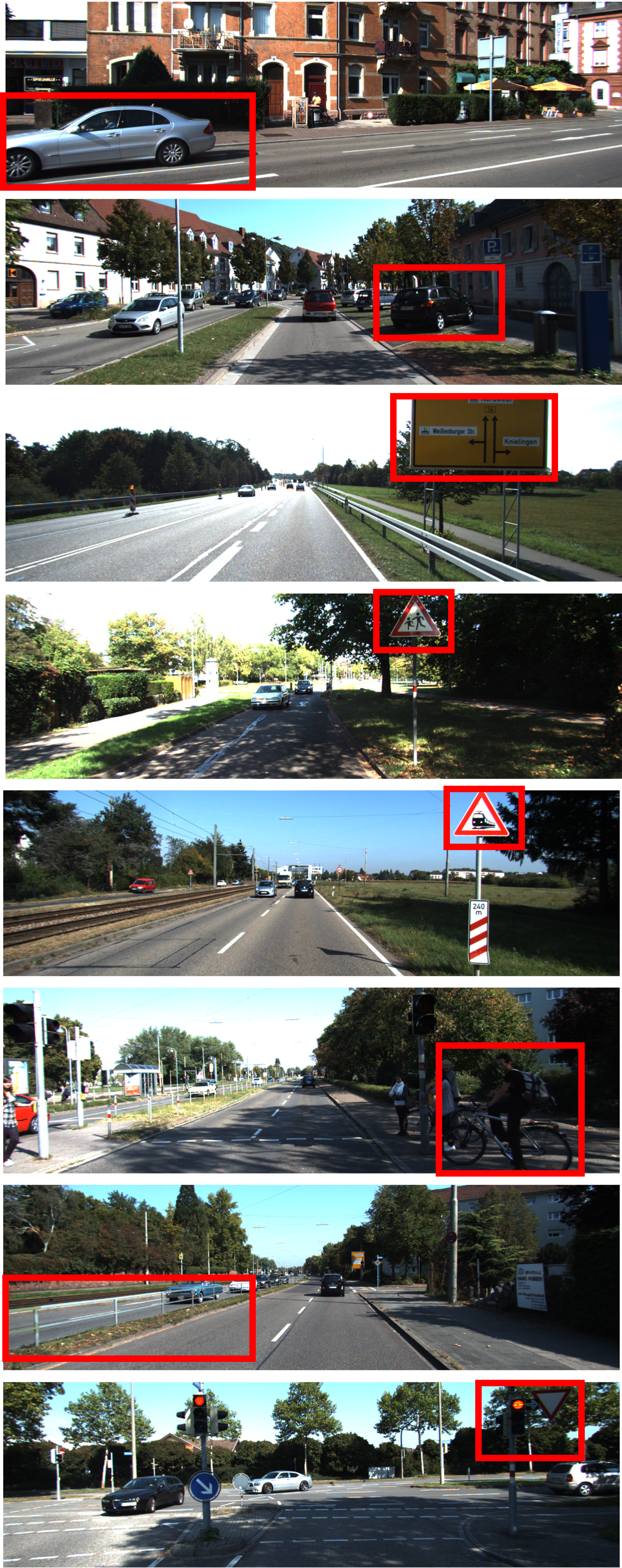}}
\hfill
\subfloat[Sceneflow]{\includegraphics[width=0.32 \linewidth]{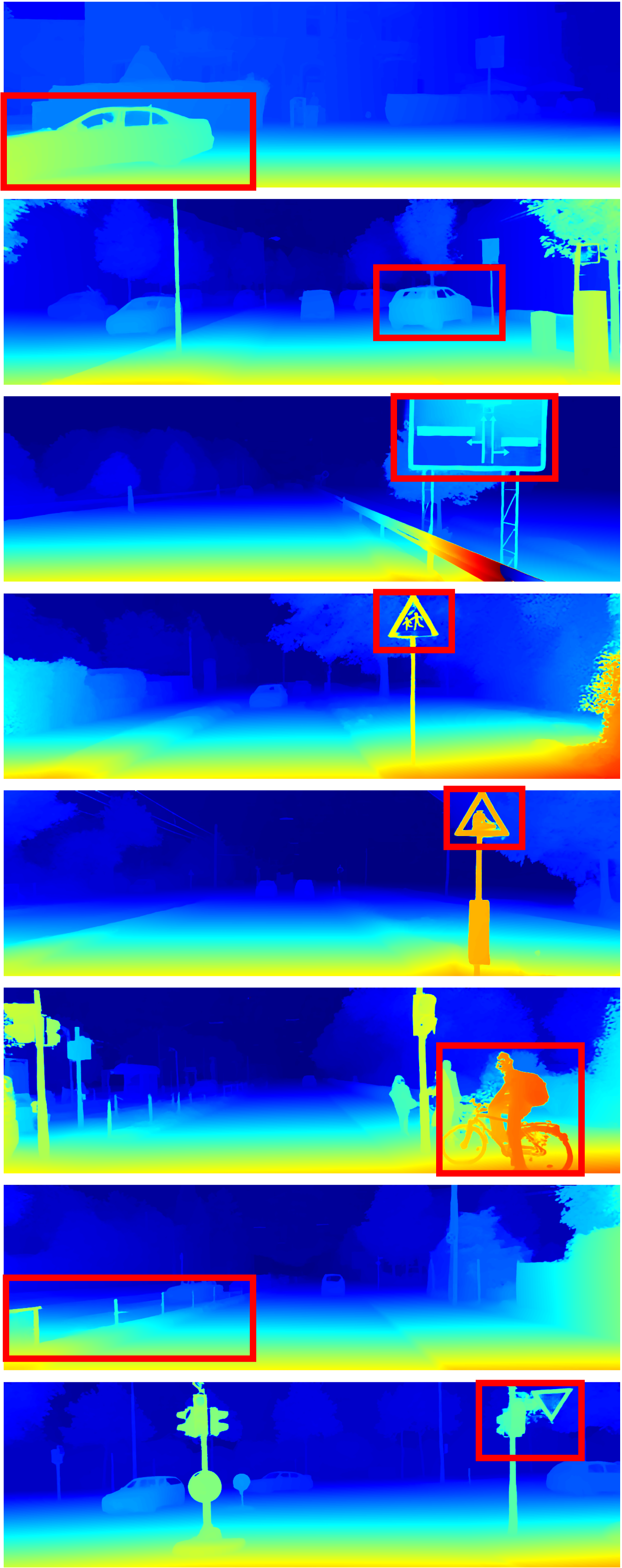}}
\hfill
\subfloat[DiffMFS]{\includegraphics[width=0.32 \linewidth]{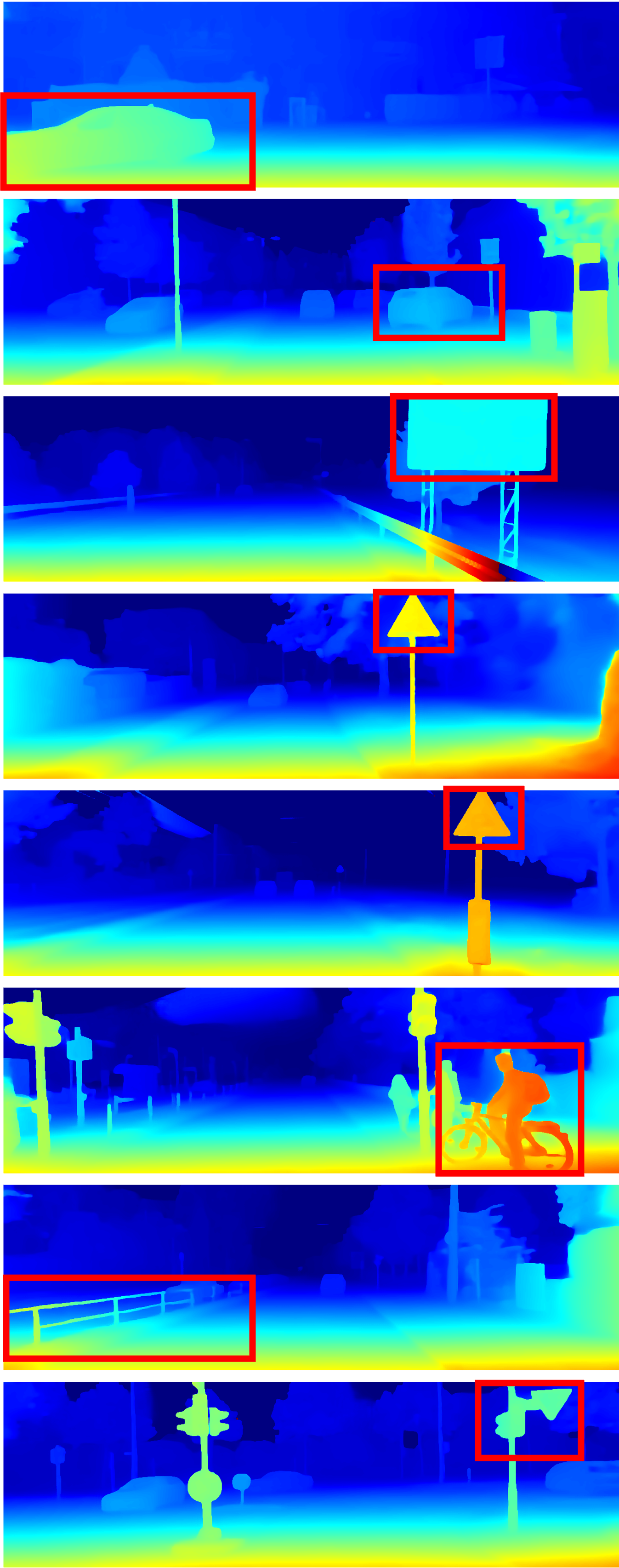}}
\caption{Zero-shot qualitative results of IGEV on KITTI dataset trained with and without our generated DiffMFS dataset. The default model is pre-traiend on sceneflow and fine-tuned by official implementation.}
\label{Zeroshot-vis}
\end{figure*}

Figure \ref{Zeroshot-vis} compares the zero-shot results of pre-training on synthetic dataset, Sceneflow and pre-training on our datasets, showing the experimental results on the ETH3D, KITTI12, and KITTI15 datasets. Although the model trained using the sceneflow method has better edge information, due to the diversity limitations of the simulated data, some object types are still missing, resulting in a decrease in the effectiveness of the algorithm. 

According to the results, we found that the model pre-trained with our generated dataset performs better on simple textures like road signs and landmarks, which shows that there are enough diverse features in our dataset for the model to learn. In addition, in the stereo matching of vehicles and pedestrians, our pre-trained model can fully identify them, while the sceneflow pre-trained model  cannot. This also proves the domain gap between the synthetic dataset and the real world.

\subsection{C2. Fine-tuned Qualitative Results}

\begin{figure*}[ht]
\centering
\subfloat[Left-view Image]{\includegraphics[width=0.32 \linewidth]{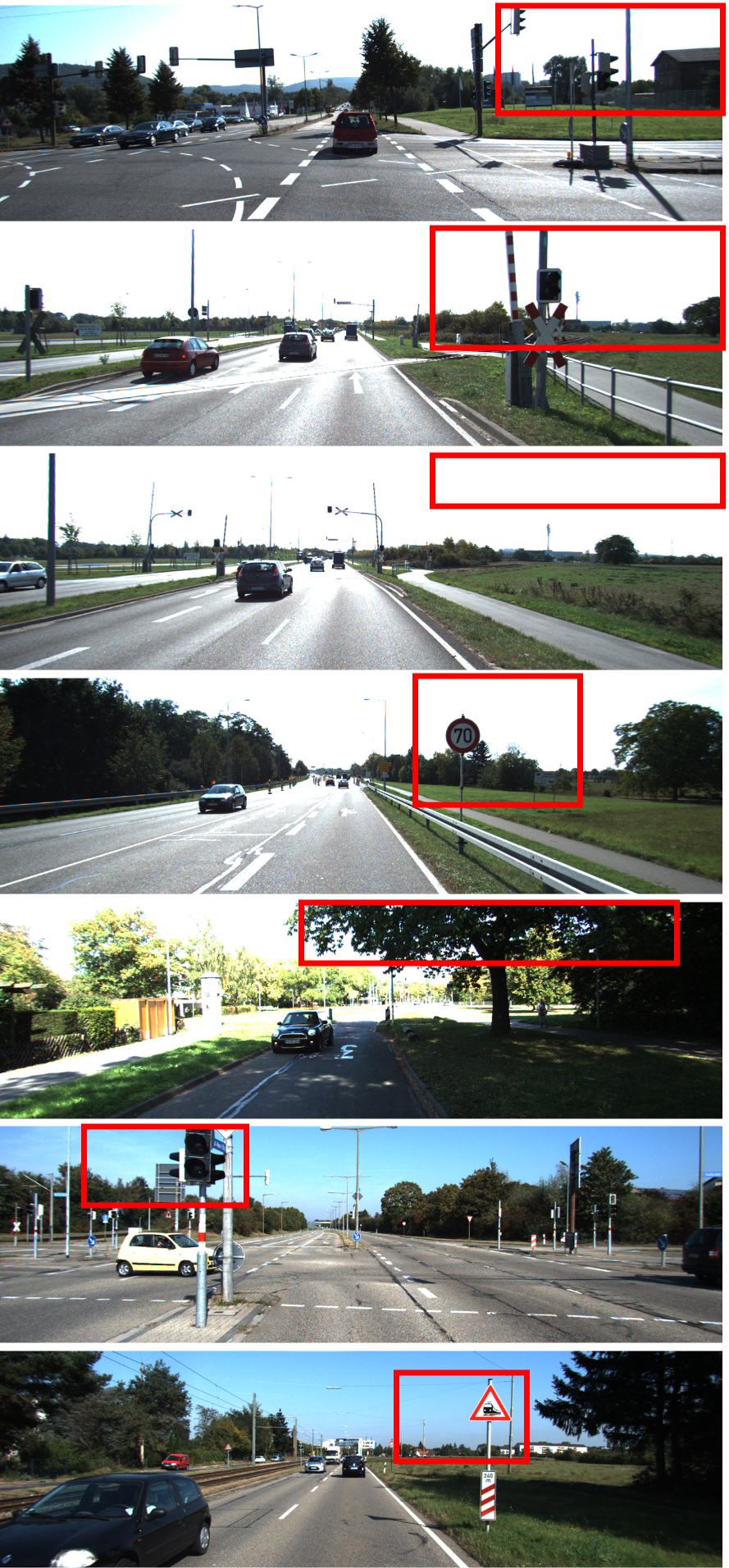}}
\hfill
\subfloat[IGEV]{\includegraphics[width=0.32 \linewidth]{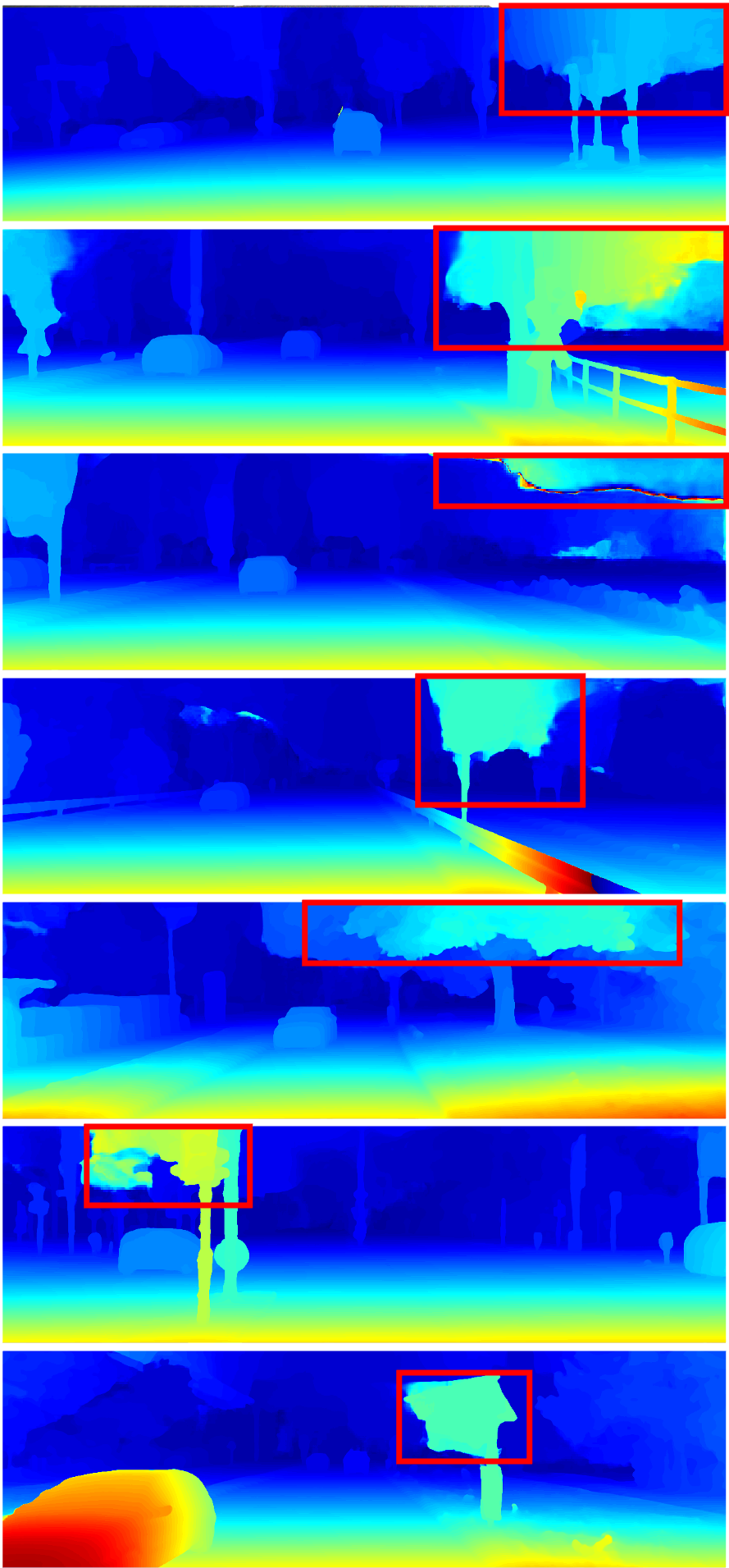}}
\hfill
\subfloat[IGEV+Ours]{\includegraphics[width=0.32 \linewidth]{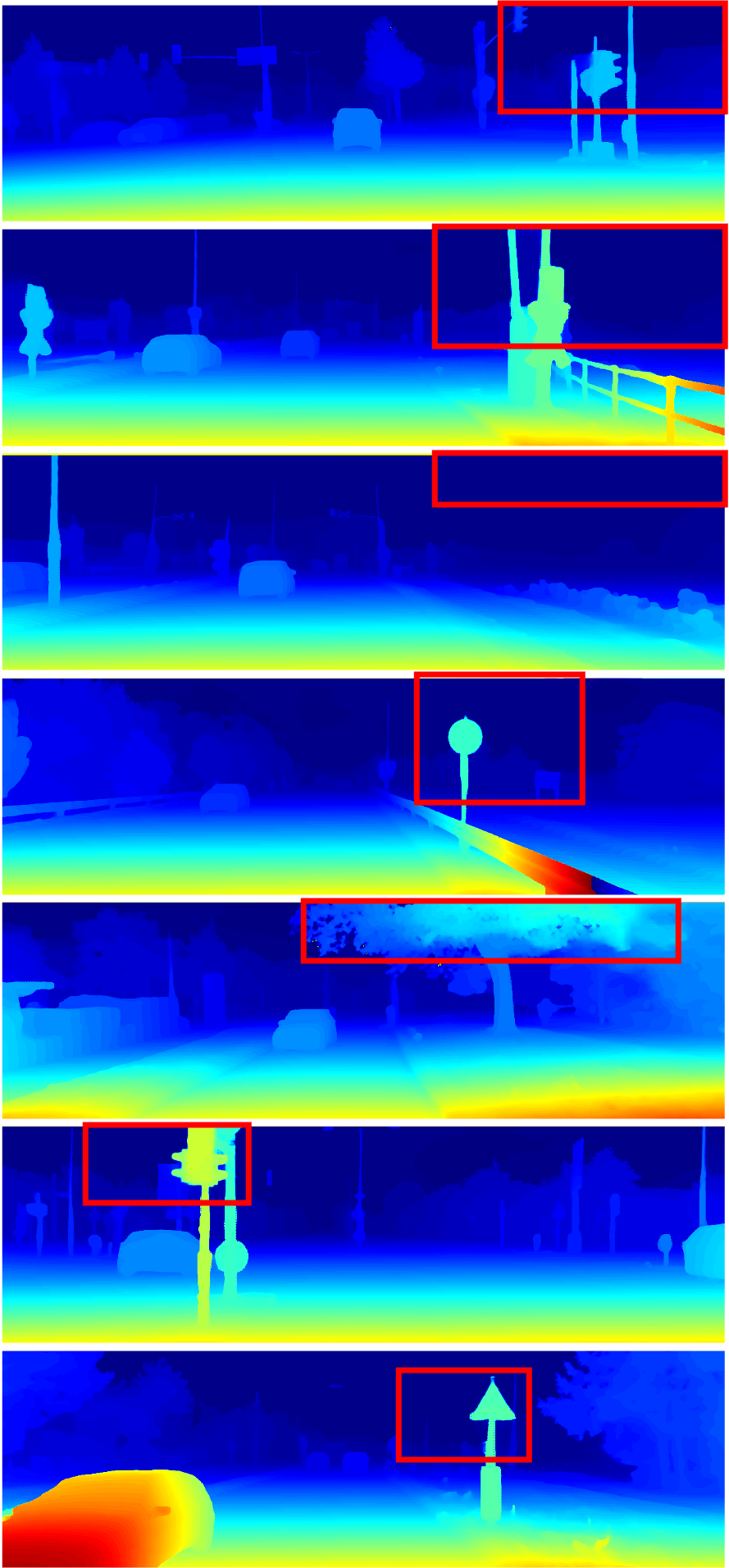}}
\caption{Qualitative results of IGEV on KITTI dataset finetuned with and without our proposed S2DKD strategy. The default model is pre-traiend on sceneflow and fine-tuned on the KITTI training set by official implementation.}
\label{Finetune-vis}
\end{figure*}

In Figure \ref{Finetune-vis}, we mainly show the effect of fine-tuning after model pre-training, and compare the indomain capabilities of different methods. We conducted  the comparison of the fine-tuning effect using simple sparse label supervision and the fine-tuning effect using our knowledge distillation method. Consistent with the superior performance on the KITTI testset, our fine-tuning method is able to generate more accurate disparity for distant objects in the sky, while traditional fine-tuning methods perform poorly due to the lack of supervisory signals in those regions.

\section{D. Limitations}

Since we use a diffusion model in the inpainting process, the data synthesis speed is slow. Our next work will focus on further improving the speed and quality of the data.

\bibliography{aaai25}